\documentclass[10pt,twocolumn,letterpaper]{article}

\usepackage{wacv}              %

\usepackage{graphicx}
\usepackage{multirow}
\usepackage{enumitem}
\usepackage[dvipsnames]{xcolor}
\usepackage{amsmath}
\usepackage{amssymb}
\usepackage{booktabs}
\usepackage{microtype}
\usepackage{indentfirst}

\usepackage[pagebackref,breaklinks,colorlinks]{hyperref}

\newcommand{\modelname}{PolyMaX\xspace}

\makeatletter
\DeclareRobustCommand\onedot{\futurelet\@let@token\@onedot}
\def\@onedot{\ifx\@let@token.\else.\null\fi\xspace}
\def\eg{\emph{e.g}\onedot} 
\def\ie{\emph{i.e}\onedot} 
 
 \def\vs{\emph{vs}\onedot}
 
\def\etal{\emph{et al}\onedot}
\makeatother

\def\wacvPaperID{1508} %
\def\confName{WACV}
\def\confYear{2024}

\usepackage[capitalize]{cleveref}
\crefname{section}{Sec.}{Secs.}
\Crefname{section}{Section}{Sections}
\Crefname{table}{Table}{Tables}
\crefname{table}{Tab.}{Tabs.}

\newcommand{\figref}[1]{Fig\onedot~\ref{#1}}
\newcommand{\equref}[1]{Eq\onedot~\eqref{#1}}
\newcommand{\secref}[1]{Sec\onedot~\ref{#1}}
\newcommand{\tabref}[1]{Tab\onedot~\ref{#1}}

\makeatletter
\long\def\comment#1{}
\renewcommand\section{\@startsection {section}{1}{\z@}%
                                   {-.8ex \@plus -2ex \@minus -.2ex}%
                                   {.5ex \@plus.2ex}%
                                   {\normalfont\large\bfseries\raggedright}}
\renewcommand\subsection{\@startsection{subsection}{2}{\z@}%
                                   {-.6ex \@plus -2ex \@minus -.2ex}%
                                   {.5ex \@plus.2ex}%
{\normalfont\large\bfseries\raggedright}}
\renewcommand\subsubsection{\@startsection{subsubsection}{3}{\z@}%
                                     {-.5ex\@plus -.2ex \@minus -.2ex}%
                                     {.2ex \@plus .2ex}%
                                     {\normalfont\large\bfseries\raggedright}}

  \renewenvironment{thebibliography}[1]{%
    \begin{oldthebibliography}{#1}%
      \small
      \setlength{\parskip}{0pt}%
      \setlength{\itemsep}{0pt}%
  }%
  {%
    \end{oldthebibliography}%
  }

\renewcommand\normalsize{%
   \@setfontsize\normalsize\@xpt\@xiipt
   \abovedisplayskip 4\p@ \@plus2\p@ \@minus5\p@
   \abovedisplayshortskip \z@ \@plus3\p@
   \belowdisplayshortskip 2\p@ \@plus3\p@ \@minus3\p@
   \belowdisplayskip \abovedisplayskip
   \let\@listi\@listI}
\renewcommand\small{%
   \@setfontsize\small\@ixpt{11}%
   \abovedisplayskip 2.5\p@ \@plus3\p@ \@minus4\p@
   \abovedisplayshortskip \z@ \@plus2\p@
   \belowdisplayshortskip 2\p@ \@plus2\p@ \@minus2\p@
   \def\@listi{\leftmargin\leftmargini
               \topsep 4\p@ \@plus2\p@ \@minus2\p@
               \parsep 2\p@ \@plus\p@ \@minus\p@
               \itemsep \parsep}%
   \belowdisplayskip \abovedisplayskip
}

\def\tightmath{
\abovedisplayskip=3pt plus 2pt minus 1pt
\abovedisplayshortskip=0pt plus 1pt minus 1pt
\belowdisplayskip=3pt plus 2pt minus 1pt
\belowdisplayshortskip=0pt plus 1pt minus 1pt }
\def\crushmath{
\abovedisplayskip=1pt plus 1pt minus 2pt
\abovedisplayshortskip=1pt plus 1pt minus 2pt
\belowdisplayskip=1pt plus 1pt minus 2pt
\belowdisplayshortskip=1pt plus 1pt minus 2pt }
\tightmath
\crushmath

\makeatother

\begin{document}

\title{\modelname: General Dense Prediction with Mask Transformer}

\author{%
  \begin{tabular}{c c c c c}
  Xuan Yang\thanks{Corresponding to \texttt{xuanyang@google.com}.}  & Liangzhe Yuan & Kimberly Wilber & Astuti Sharma & Xiuye Gu \\
  \multicolumn{5}{c}{Siyuan Qiao  \ \ \ Stephanie Debats \ \ \ Huisheng Wang \ \ \ Hartwig Adam} \\
  \multicolumn{5}{c}{Mikhail Sirotenko \ \ \ \ \ Liang-Chieh Chen\thanks{Work done while at Google. Now at ByteDance.}} \\
  \multicolumn{5}{c}{Google Research} \\
  \end{tabular}
  }

\maketitle

\begin{abstract}
 Dense prediction tasks, such as semantic segmentation, depth estimation, and surface normal prediction, can be easily formulated as per-pixel classification (discrete outputs) or regression (continuous outputs).
 This per-pixel prediction paradigm has remained popular due to the prevalence of fully convolutional networks.
 However, on the recent frontier of segmentation task, the community has been witnessing a shift of paradigm from per-pixel prediction to cluster-prediction with the emergence of transformer architectures, particularly the mask transformers, which directly predicts a label for a mask instead of a pixel.
 Despite this shift, methods based on the per-pixel prediction paradigm still dominate the benchmarks on the other dense prediction tasks that require continuous outputs, such as depth estimation and surface normal prediction.
 Motivated by the success of DORN and AdaBins in depth estimation, achieved by discretizing the continuous output space, we propose to generalize the cluster-prediction based method to general dense prediction tasks. This allows us to unify dense prediction tasks with the mask transformer framework.
 Remarkably, the resulting model \modelname demonstrates state-of-the-art performance on three benchmarks of NYUD-v2 dataset. We hope our simple yet effective design can inspire more research on exploiting mask transformers for more dense prediction tasks.
 Code and model will be made available\footnote{https://github.com/google-research/deeplab2}.
\end{abstract}

\section{Introduction}
\label{sec:intro}

Entering the deep learning era~\cite{krizhevsky2012imagenet,simonyan2015very,he2016deep}, enormous efforts have been made to tackle dense prediction problems,
including but not limited to,
image segmentation~\cite{long2014fully,deeplabv12015,chen2018deeplabv2,chen2017deeplabv3,zhao2017pyramid}, depth estimation~\cite{SilbermanECCV12,eigen2014depth,fu2018deep,lee2019big,Ranftl2021,Ranftl2022}, surface normal prediction~\cite{eigen2015predicting,wang2015designing,misra2016cross,kokkinos2017ubernet}, and others~\cite{amfm_pami2011,mottaghi2014role,lin2014microsoft,dosovitskiy2015flownet,Cordts2016Cityscapes,taskonomy2018}.
Early attempts formulate these problems as per-pixel prediction (\ie, assigning a predicted value to every pixel) via fully convolutional networks~\cite{long2014fully}.
Specifically, when the desired prediction of each pixel is discrete, such as image segmentation, the task is constructed as per-pixel \textit{classification}, while other tasks whose target outputs are continuous, such as depth and surface normal, are instead cast as per-pixel \textit{regression} problems.

\begin{figure}[t]
\includegraphics[width=0.99\linewidth]{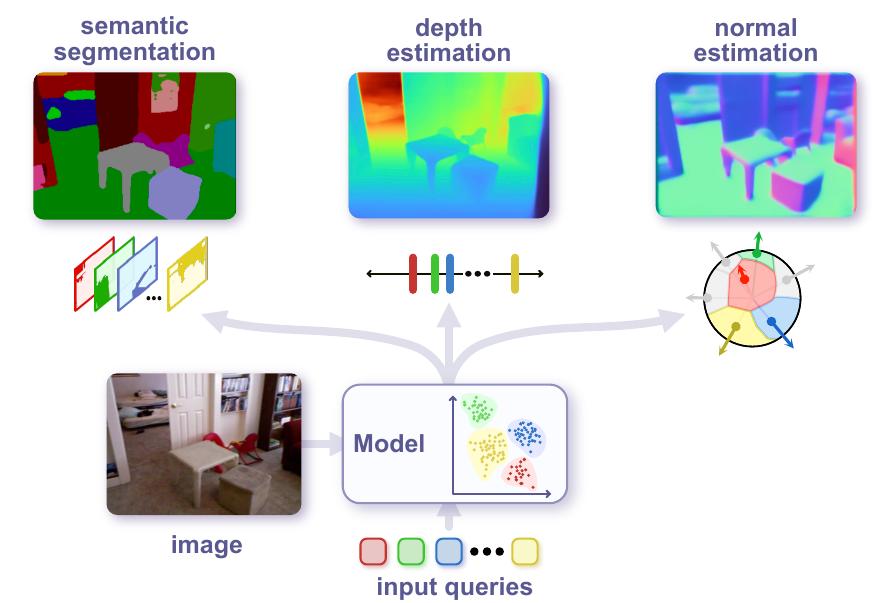}
\vspace*{-2mm}
\caption{\label{fig:teaser}
\textbf{Cluster-prediction forms the foundation of our unified \textit{Mask-Transformer}-based framework \modelname for dense prediction tasks.}
In the cluster-prediction paradigm, the model learns to transform input queries to cluster centers, where each cluster center learns to group similar pixels together. 
Pixels of the same group are assigned to the same label.
This paradigm works well for discrete output space, such as segmentation.
To extend it for continuous and even high-dimensional output space (\eg, depth estimation and surface normal), we dynamically partition the output space into clusters, dependent on the input image.
}
\end{figure}

Recently, a new paradigm for segmentation tasks is drawing attention because of its superior performance compared to previous per-pixel classification approaches. Inspired by the object detection network DETR~\cite{carion2020end}, MaX-DeepLab~\cite{maxdeeplab} and MaskFormer~\cite{cheng2021per} propose to classify each segmentation mask as a whole instead of pixel-wise, by extending the concept of object queries in DETR to represent the clusters of pixels in segmentation masks.
Specifically, with the help of pixel clustering via conditional convolutions~\cite{jia2016dynamic,tian2020conditional,wang2020solov2}, these \textit{Mask-Transformer}-based works employ transformer decoders~\cite{transformer} to convert object queries to a set of (mask embedding vector, class prediction) pairs, which finally yield a set of binary masks by multiplying the mask embedding vectors with the pixel features.
Such methods are effective when the target domain is \textit{discretely} quantized (\eg, one semantic label is encoded by one integer scalar, as in image segmentation). However, it is unclear how this framework can be generalized to other dense prediction tasks, whose outputs are \emph{continuous} or even multi-dimensional, such as depth estimation and surface normal prediction.

On the contrary, in the field of depth estimation, DORN~\cite{fu2018deep} demonstrates the potential of \textit{discretizing} the continuous depth range into a set of fixed bins, and AdaBins~\cite{bhat2021adabins} further adaptively estimates the bin centers, dependent on the input image.
The continuous depth values are then estimated by linearly combining the bin centers.
Promising results are achieved by jointly learning the bin centers and performing per-pixel \textit{classification} on those bins.
This insight -- of using classification to perform dense prediction in a continuous domain -- opens up the possibility of performing many other continuous dense prediction tasks within the Mask-Transformer-based framework, a powerful tool for discrete value predictions.

Consequently, a few natural questions emerge: \textit{Can we extend these Mask-Transformer-based frameworks even further, to solve more continuous dense prediction tasks}?
\textit{Can the resulting framework generalize to other multi-dimensional continuous domain, \eg, surface normal estimation?}
In this work, we provide affirmative answers to those questions by proposing a new general architecture for dense prediction tasks.
Specifically, building on top of the insight from~\cite{fu2018deep,bhat2021adabins}, we generalize the mask transformer framework~\cite{yu2022kmax} to multiple dense prediction tasks by
using the cluster centers (\ie, object queries) as the intermediate representation. 
We evaluate the resulting model, called \textbf{\modelname}, on the challenging NYUD-v2~\cite{SilbermanECCV12} and Taskonomy~\cite{taskonomy2018} datasets.
Remarkably, our simple yet effective approach demonstrates new state-of-the-art performance on semantic segmentation, depth estimation, and surface normal prediction, without using any extra modality as inputs (\eg, multi-modal RGB-D inputs as in CMNeXt~\cite{zhang2023delivering}), heavily pretrained backbones (\eg, Stable-Diffusion~\cite{rombach2021highresolution} as in VPD~\cite{zhao2023unleashing}) or complex pretraining schemes (\eg, a mix of 12 datasets as in ZoeDepth~\cite{bhat2023zoedepth}).

Our contributions are summarized as follows:

\begin{itemize}
    \item We propose \modelname, which generalizes dense prediction tasks with a unified \textit{Mask-Transformer}-based framework. We take surface normal as a concrete example to demonstrate how general dense prediction tasks can be solved by the proposed framework.
    \item We evaluate \modelname on NYUD-v2 and Taskonomy datasets, and it sets new state-of-the-arts on multiple benchmarks on NYUD-v2, achieving 58.08 mIoU, 0.250 root-mean-square (RMS) error and 13.09 mean error on semantic segmentation, depth estimation and surface normal prediction, respectively. %
    \item We further perform the scalability study, which demonstrates that \modelname scales significantly better than the conventional per-pixel regression based methods as the pretraining data increases. 
    \item Lastly, we provide the high-quality pseudo-labels of semantic segmentation for Taskonomy dataset, aiming to compensate for the scarcity of existing large-scale multi-task datasets and facilitate the future research. 
\end{itemize}
\section{Related Work}
\label{sec:related}

\begin{figure*}[t]
\includegraphics[width=0.99\linewidth]{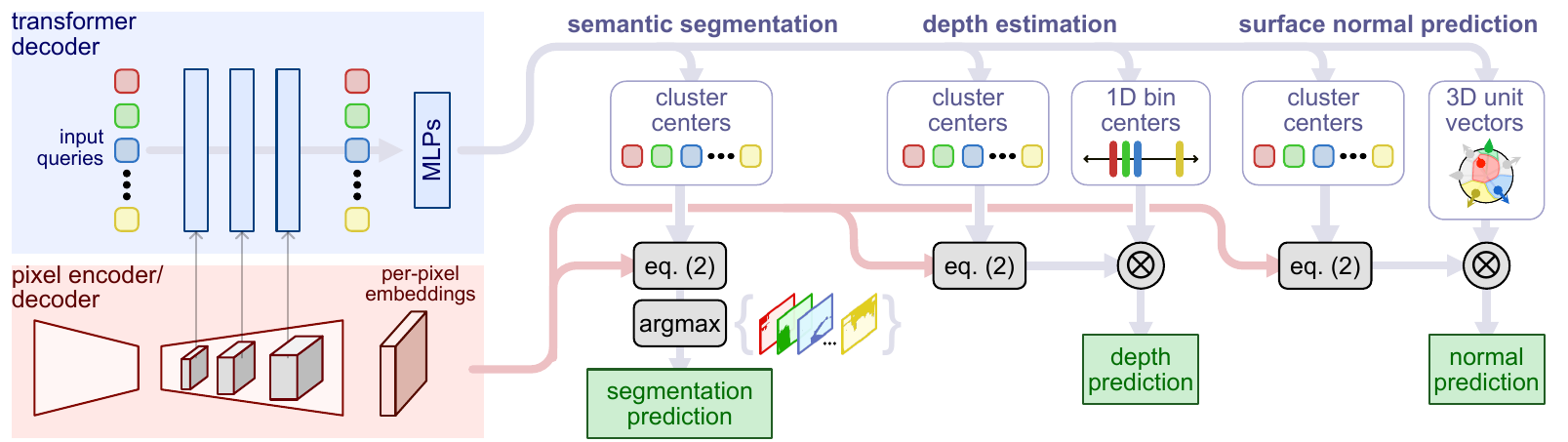}
\vspace*{-2mm}
\caption{
\textbf{The proposed model \modelname unifies dense prediction tasks with the mask transformer framework, where cluster centers are used as an intermediate representation.}
The mask transformer framework contains two paths: (1) pixel encoder/decoder to generate per-pixel embeddings, and (2) transformer decoder to generate cluster centers (\ie, object queries) from input queries.
As shown in~\equref{eq:dense_prediction}, the probability distribution map is generated by first multiplying the per-pixel embeddings with the cluster centers, followed by a softmax operation.
For semantic segmentation, the final segmentation map is obtained by applying the argmax operation to find the maximum probability semantic label.
For depth estimation, we further apply another MLP (omitted in the Figure) to estimate the bin center values, which are then linearly combined with the probability distribution map to form the final depth map.
Similarly, for surface normal prediction, the estimated 3D unit vectors are linearly combined with the probability distribution map to form the final surface normal map.
$ \bigotimes $ denotes the linear combination operation.
}
\label{fig:explanation}
\end{figure*}

\textbf{Dense Prediction in the Discrete Domain}\quad
Image segmentation partitions images into multiple segments by their semantic classes.
For example, fully convolutional networks~\cite{long2015fully} train semantic segmentation in an end-to-end manner mapping pixels into their classes~\cite{liu2015parsenet,li2017fully,yuan2017automatic,liu2016accurate}.
Atrous convolution increases the network receptive field for semantic segmentation without additional learnable parameters~\cite{chen2017deeplabv3,deeplabv12015,chen2018deeplabv2,wang2018understanding,yang2018denseaspp,deeplabv3plus2018,qiao2021vip}.
Many state-of-the-art methods use an encoder-decoder meta architecture for stronger global and local information integration~\cite{noh2015learning,kendall2015bayesian,chen2016attention,badrinarayanan2017segnet,fu2019stacked,cheng2019panoptic,yuan2020object,wang2020axial,gu2023dataseg,sun2023remax,yu2023convolutions}.

\textbf{Dense Prediction in the Continuous Domain}\quad 
Unlike image segmentation that produces discrete predictions, depth and surface normal estimation expect the models to predict continuous values~\cite{eigen2014depth,qi2018geonet}. Most early attempts~\cite{laina2016deeper,liu2015learning,xu2018structured,qiao2021vip} solve it as a standard regression problem. 
To alleviate training instability, DORN~\cite{fu2018deep} proposes treating the problem as classification by discretizing the continuous depth into a set of pre-defined intervals (bins). AdaBins~\cite{bhat2021adabins} adaptively learns the depth bins, conditioned on the input samples. LocalBins~\cite{bhat2022localbins} further learns the depth range partitions from local regions instead of global distribution of depth ranges.
BinsFormer~\cite{li2022binsformer} views adaptive bins generation as a direct set prediction problem~\cite{detr}.
The current state-of-art on depth estimation is set by VPD~\cite{zhao2023unleashing}, which employs the Stable-Diffusion~\cite{rombach2021highresolution} pretrained with the large-scale LAION-5B~\cite{schuhmann2022laion} dataset.
Among the non-diffusion-based models, ZoeDepth~\cite{bhat2023zoedepth} shows the strongest capability by pretraining on 12 datasets with relative depth, and then finetuning on two datasets with metric depth. 

\textbf{Surface Normal Prediction}\quad 
Despite having a continuous output domain like depth estimation, the surface normal problem remains under-explored for discretization approaches. Most prior works on surface normal estimation focus on improving the loss~\cite{bae2021estimating,liao2019spherical,do2020surface}, reducing the distribution bias~\cite{bae2021estimating} and shift~\cite{do2020surface}, and resolving the artifacts in ground-truth by leveraging other modalities~\cite{hickson2019floors, qi2020geonet++}. Until recently, iDisc~\cite{piccinelli2023idisc}, a method based on discretizing internal representations, shows promise for depth estimation and surface normal prediction. However, it differs from depth estimation discretization methods by enforcing discretization in the \emph{feature} space rather than the target output space. We are more interested in the latter approach, as it is more generalizable to other dense prediction tasks.

\textbf{Other Efforts to Unify Dense Prediction Tasks}\quad 
Realizing segmentation, depth and surface normal are all pixel-wise mapping problem, previous works~\cite{wang2023images,long2021adaptive,li2022learning, ning2023all, bhattacharjee2023vision} have  deployed them to the same framework. Some works~\cite{bruggemann2021exploring, ye2022invpt, xu10multi, xu2022mtformer, bhattacharjee2022mult}  improve performance by exploring the relations among dense prediction tasks. UViM~\cite{kolesnikov2022uvim} and Painter~\cite{wang2023images} unify segmentation and depth estimation, but neither of them is based on discretizing the continuous output space, and neither considers surface normal estimation. By contrast, we focus on a complementary perspective: a unified architecture for dense prediction that models both discrete and continuous tasks, covering image segmentation, depth estimation, and surface normal prediction.

\textbf{Mask Transformer}\quad
Witnessing the success of transformers~\cite{transformer} in NLP, the vision community has begun exploring them for more computer vision tasks~\cite{dosovitskiy2020image}.
Inspired by DETR~\cite{detr} and conditional convolutions~\cite{jia2016dynamic}, MaX-DeepLab~\cite{maxdeeplab} and MaskFormer~\cite{cheng2021per} switch the segmentation paradigm from per-pixel classification to mask classification, by introducing the \textit{mask transformer}, where each input query learns to correspond to a mask prediction together with a class prediction.
Several works~\cite{zheng2021rethinking,strudel2021segmenter,xie2021segformer} also introduce transformer into semantic segmentation. Further improvements are made on attention mechanism to enhance the performance of mask transformers~\cite{zhu2020deformable,panoptic_segFormer,cheng2021mask2former}.
CMT-DeepLab~\cite{yu2022cmt}, kMaX-DeepLab~\cite{yu2022kmax}, and ClustSeg~\cite{liang2023clustseg} reformulate the cross-attention learning in transformer as a clustering process.
Our proposed \modelname builds on top of the cluster-based mask transformer architecture~\cite{yu2022kmax}.
Rather than focusing on segmentation tasks, \modelname unifies dense prediction tasks (\ie, image segmentation, depth estimation, and surface normal estimation) by extending the cluster-based mask transformer to support both discrete and continuous outputs.

\section{Method}
\label{sec:method}

In this section, we first describe how segmentation~\cite{yu2022kmax} and depth estimation~\cite{bhat2021adabins} can be transformed from per-pixel prediction problems to cluster-prediction problems (\secref{sec:seg_depth}).
We then introduce our proposed method, \modelname, a new mask transformer framework for general dense predictions (\secref{sec:polymax}).
We take surface normal prediction as a concrete example to explain how general dense prediction problems can be reformulated in a similar fashion, allowing us to unify them into the same mask transformer framework.

\subsection{Mask-Transformer-Based Segmentation and Depth Estimation}
\label{sec:seg_depth}

\textbf{Cluster-Prediction Paradigm for Segmentation}\quad
Two earlier works, MaX-DeepLab~\cite{maxdeeplab} and MaskFormer~\cite{cheng2021per}, demonstrate how to shift image segmentation from per-pixel classification to the cluster-prediction paradigm.
Along the same direction, we follow the recently proposed clustering perspective~\cite{yu2022cmt,yu2022kmax}, which casts object queries to cluster centers. This paradigm is realized by the following two steps:
\begin{enumerate}%
    \item \textit{pixel-clustering}: group pixels into $K$ clusters, represented by segmentation masks $\{ \mathbf{m_i} | \mathbf{m_i} \in [0, 1]^{H \times W} \}^K_{i=1}$, where $H$ and $W$ are height and width. Note that $\mathbf{m_i}$ denotes soft segmentation masks.
    \item \textit{cluster-classification}: assign semantic label to each cluster with a probability distribution over $C$ classes. The $i$th cluster's probability distribution is a 1D vector, denoted as $\mathbf{p_i}$, where $\{ \mathbf{p_i} | \mathbf{p_i} \in [0, 1]^C, \sum_{c=1}^C \mathbf{p}_{i,c}=1\}^K_{i=1}$.
\end{enumerate}

These clusters and probability distributions are jointly learned to predict the output $S$, a set of $K$ cluster-probability pairs: 
\begin{equation}
\label{eq:output_s}
    S = \{ ( \mathbf{m_i}, \mathbf{p_i} ) \}^K_{i=1}
\end{equation}
This cluster-prediction paradigm is general for semantic~\cite{he2004multiscale}, instance~\cite{hariharan2014simultaneous}, and panoptic~\cite{kirillov2018panoptic} segmentation.
When only handling semantic segmentation, the framework can be further simplified by setting $K=C$ (\ie, number of clusters is equal to number of classes) which has a fixed matching between cluster centers (\ie, object queries) and semantic classes. We adopt this simplification, since the datasets we experimented with only support semantic segmentation.

\textbf{Cluster-Prediction Paradigm for Depth Estimation}\quad
At first glance, depth estimation seems incompatible with this cluster-prediction paradigm because of its continuous nature.
However, recent works~\cite{fu2018deep,bhat2021adabins} propose promising solutions by dividing the continuous depth range into $K$ learnable bins and regarding the task as a classification problem.
By this means, the depth estimation task fits neatly into the above cluster-classification paradigm.
Specifically, the \textit{pixel-clustering} step outputs the range attention map $\{\mathbf{r_i} | \mathbf{r_i} \in [0, 1]^{H\times W}\}_{i=1}^K$, which (after softmax) represents the predicted probability distribution over the $K$ bins for each pixel. 
While the \textit{cluster-prediction} step estimates bin centers $\{\mathbf{b_i}\}_{i=1}^K$, adaptively discretizing the continuous depth range for each image.
Here, $K$ controls the granularity of the depth range partition. Therefore, the output of depth estimation can be expressed as $ \{ ( \mathbf{r_i}, \mathbf{b_i} ) \}^K_{i=1} $, sharing the same formulation as segmentation (Eq~\ref{eq:output_s}). 
The final depth prediction is then generated by a linear combination between the depth values of bin centers and the range attention map.

\begin{figure}[t!]
\includegraphics[width=0.99\linewidth]{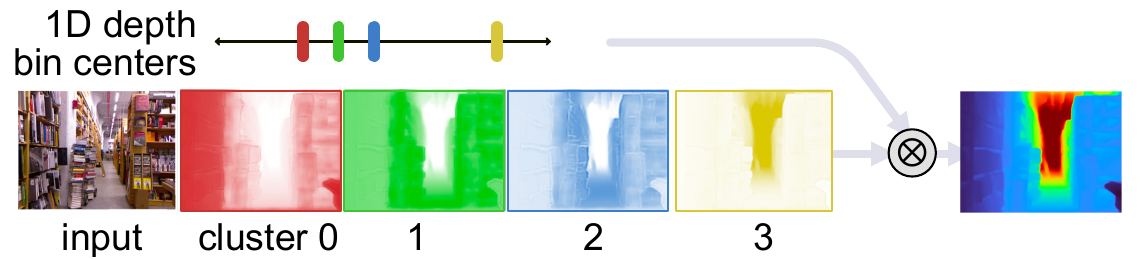}
\label{fig:depth-explain}
\vspace*{-7mm}
\caption{\textbf{For the task of depth prediction, we illustrate one example with four depth bins (\ie, $K=4$).} 
The probability distribution map (\ie, output of~\equref{eq:dense_prediction}) has shape $HW\times 4$ (generated by four cluster centers), which represents the predicted probability distribution over 4 bins for each pixel. As shown in the figure, the first cluster center (red) learns to highlight closest pixels; the second and third clusters (green/blue) highlight mid-range pixels, and the fourth (yellow) cluster highlights the furthest pixels.
These maps are linearly combined (denoted as $\bigotimes$) with their
corresponding 1D bin centers to produce the final depth map prediction.
}
\end{figure}

\textbf{Mask Transformer Framework}\quad
After describing the cluster-prediction paradigm for both segmentation and depth estimation, we now explain how to integrate them to the mask transformer framework.
In the framework, the $K$ cluster centers are learned from the $K$ input queries through the transformer-decoder branch (blue block in Fig.~\ref{fig:explanation}). It associates the clusters, represented as query embedding vectors, with the pixel features extracted from the pixel encoder-decoder path~\cite{ronneberger2015u} (pink block in Fig.~\ref{fig:explanation}) through self-attention and cross-attention blocks.
The aggregated information is gradually refined as the pixel features go from lower-resolution to higher-resolution.
The \textit{probability distribution map} is then generated from the following equation:
\begin{equation}
    \label{eq:dense_prediction}
    \operatornamewithlimits{softmax}_{K} (\mathbf{F} \times \mathbf{Q}^T),
\end{equation}
 where $\mathbf{F} \in \mathbb{R}^{HW \times D}$ and $\mathbf{Q} \in \mathbb{R}^{K \times D}$ denote the final pixel features (\ie, per-pixel embeddings) and cluster centers (\ie, object queries). $D$ is the channel dimension of pixel features and queries.   
For segmentation, the probability distribution map corresponds to the segmentation masks $\{\mathbf{m_i}\}$, while for depth estimation, it becomes the range attention map $\{\mathbf{r_i}\}$.
Finally, another Multi-Layer Perceptrons (MLPs) are added on top of the cluster centers $\mathbf{Q}$ to predict the probability distribution $\mathbf{p_i}$ or bin centers $\mathbf{b_i}$ for segmentation and depth estimation, respectively.
Consequently, the cluster centers in the mask transformer are used as an intermediate representation in both segmentation and depth estimation.

\subsection{Mask-Transformer-Based General Dense Prediction}
\label{sec:polymax}
General dense prediction tasks can be expressed as follows:
\begin{equation}
 \mathcal{T}: \mathbb{R}^{H \times W \times 3} \to \mathbb{R}^{H \times W \times D_\mathcal{T}} \\
   , D_\mathcal{T} \in \mathbb{N}
\end{equation}
As presented in the above formulation, the objective of general dense prediction is to learn the pixel mapping $\mathcal{T}$ from the input RGB space to the structured output space $\mathbb{R}$ with different dimensionalities ($D_\mathcal{T}$) and topologies (output space $\mathbb{R}$ is either discrete or continuous). For instance, segmentation and depth estimation task correspond to the single-dimensional ($D_\mathcal{T}=1$) output space of discrete and continuous topology, respectively (\tabref{tab:target_domain}).
Given that the cluster-prediction methods have been effectively applied to these tasks (as described in~\secref{sec:seg_depth}), we take a step further and study how general dense prediction tasks, particularly multi-dimensional continuous target space (\eg, surface normal), can be learned by the cluster-prediction approach.     

\begin{table}[t!]
\centering
\begin{tabular}{c|cc}
target domain & output space $\mathbb{R}$ & $D_\mathcal{T}$ \\
\toprule[0.12em]
segmentation & discrete & 1\\
depth estimation & continuous & 1\\
surface normal & continuous & 3\\
\end{tabular}
\vspace*{-3mm}
\caption{
\textbf{General dense prediction learns the pixel mapping from the input RGB space to the structured output space $\mathbb{R}$ with different dimensionalities $D_\mathcal{T}$.}
}
\label{tab:target_domain}
\end{table}

\textbf{Task of Surface Normal}\quad
Surface normal is a representative dense prediction task that targets a multi-dimensional continuous output space, in which case $D_\mathcal{T} = 3$ and output space $\mathbb{R}$ is continuous. This task aims at predicting, for each pixel, the direction that is perpendicular to the tangent plane of the point in a 3D world, which can be expressed by a 3D unit vector $\mathbf{v}=[N_x, N_y, N_z]$, where $N_x^2+N_y^2+N_z^2=1$.

\textbf{Cluster-Prediction for Surface Normal}\quad
From the geometric perspective, a surface normal lies on the surface of a unit 3D ball. Therefore, the problem essentially becomes learning the pixel mapping $\mathcal{T}$ from the 2D image space to the 3D point on the unit ball surface. The geometric meaning of the surface normal prediction allows us to naturally adopt the clustering-prediction approach. 

\begin{figure}[t!]
\includegraphics[width=0.99\linewidth,trim={0 0 0 0.5cm}]{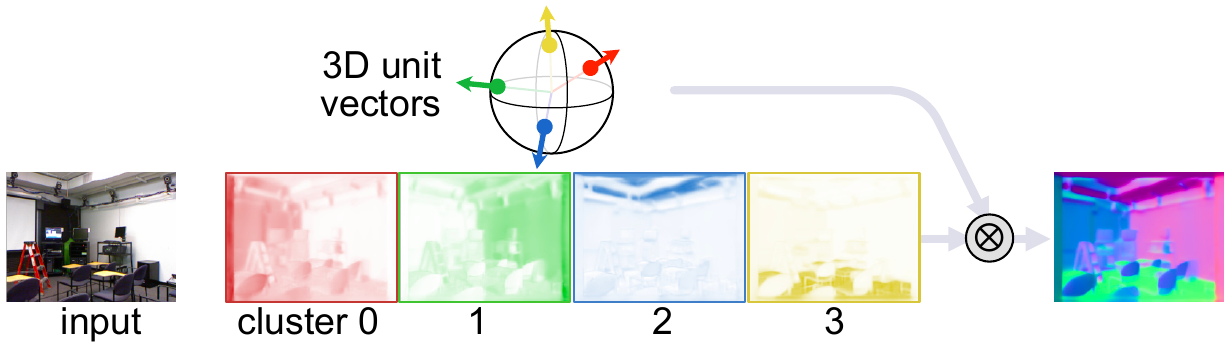}
\vspace*{-3mm}
\caption{\label{fig:normal-explanation}\textbf{To illustrate the task of surface normal prediction, we show an example model with four cluster centers} that roughly correspond to cardinal directions (3-vectors on the unit ball). Each channel of the $HW \times 4$ probability distribution map represents the predicted probability of the corresponding vector direction. These maps are linearly combined (denoted as $\bigotimes$) with their corresponding 3D unit vectors to generate the final surface normal prediction.
}
\end{figure}

Similar to~\cite{fu2018deep,bhat2021adabins},
we discretize the surface normal output space into $K$ pieces of partitioned 3D spheres, converting the problem to classification among the $K$ pieces. We refer the partitioned sphere as sphere segments for simplicity \footnote{Note that here the sphere segment does not rigorously match the mathematical definition of spherical segment, which refers to the solid produced by cutting a sphere with a pair of parallel planes.}, which can be viewed as clusters of 3D points on the unit ball surface, and thus are likely to have irregular shapes. Fortunately, with such a simplification, we can adopt the same paradigm as segmentation and depth estimation to tackle surface normals, where $\mathbf{p_i}$ can be regarded as the probability over the $K$ sphere segments. We jointly learn the center coordinates of sphere segments with the 3D-point-cluster and probability distribution. The final surface normal is predicted as a linear combination of the center coordinates of $K$ sphere segments and the 
range attention map.

\textbf{Mask Transformer Framework for General Dense Prediction}\quad
Extending the cluster-prediction paradigm to general dense prediction tasks allows us to deploy and unify them into the mask transformer framework described in \secref{sec:seg_depth}.
Going from the single-dimension ($D_\mathcal{T}=1$) to multi-dimension ($D_\mathcal{T}>1$), the major upgrade happens at the place where the cluster center $\mathbf{Q} \in \mathbb{R}^{K \times D}$ is mapped (via MLPs) to $\mathbb{R}^{K \times D_\mathcal{T}} (D_\mathcal{T} > 1)$, instead of $\mathbb{R}^{K}$, when predicting $D_\mathcal{T}$-dimensional bin centers. For instance, in the case of surface normal, each cluster center (\ie, object query) predicts a unit vector $\mathbf{v_i} \in \mathbb{R}^3$.

\textbf{Model Instantiation}\quad
We build the proposed general framework on top of kMaX-DeepLab~\cite{yu2022kmax} with
the official code-base~\cite{deeplab2_2021}. 
We call the resulting model \modelname, a \textbf{Poly}math with \textbf{Ma}sk \textbf{X}former for general dense prediction tasks.
The model architecture is illustrated in~\figref{fig:explanation}.
\section{Experimental Results}
\label{sec:results}

In this section, we first provide the details of our experimental setup.
We then report our main results on NYUD-v2~\cite{SilbermanECCV12} and Taskonomy~\cite{taskonomy2018}. We also present visualizations to obtain deeper insights of the proposed cluster-prediction paradigm, followed by ablation studies.

\subsection{Experimental Setup}

\textbf{Dataset}\quad
We focus on semantic segmentation, monocular depth estimation and surface normal tasks during experiments, as they comprehensively represent the diverse target space of dense prediction (one-dimensional to multi-dimensional, and discrete to continuous).
Specifically, we use NYUD-v2~\cite{SilbermanECCV12} dataset. The official release provides 795 training and 654 testing images with real (instead of pseudo labels) ground-truths of those three tasks.

\begin{figure}[ht!]
\centering
\newcommand{\tskfig}[1]{\includegraphics[width=0.15\linewidth]{figs/taskonomy/#1.jpeg}\hspace{2pt}}
\begin{tabular}{r@{\hspace{2pt}}l}
\rotatebox{90}{\hspace{10pt}\footnotesize input}&%
\tskfig{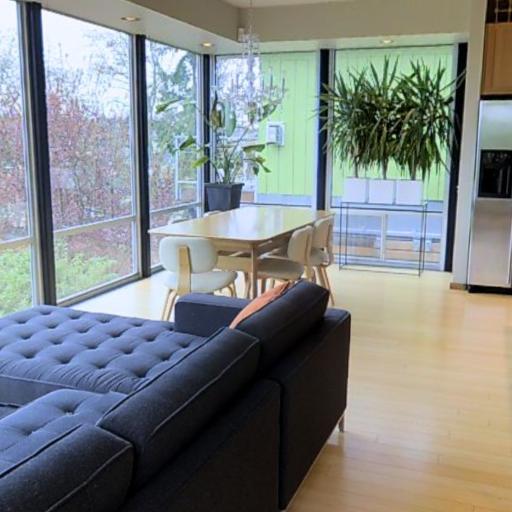}\tskfig{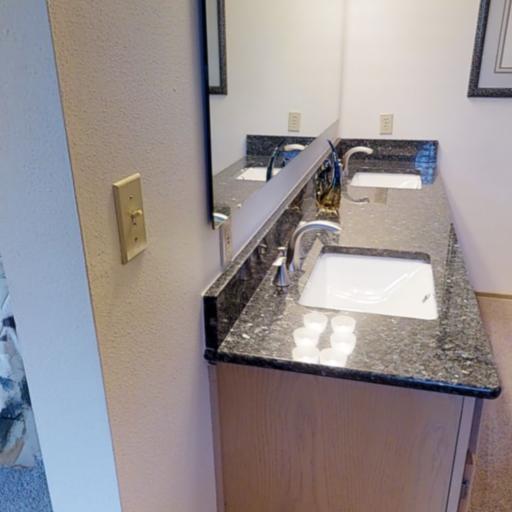}\tskfig{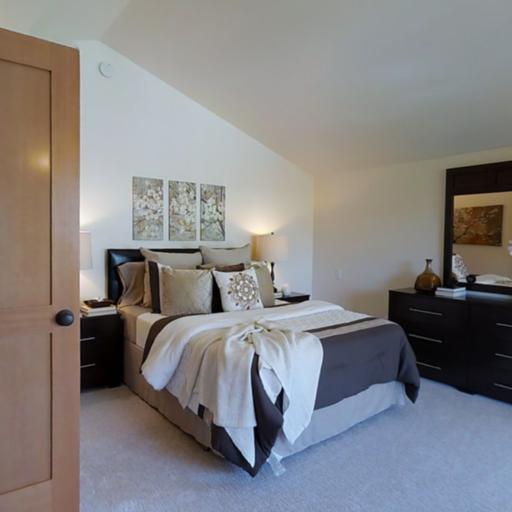}\tskfig{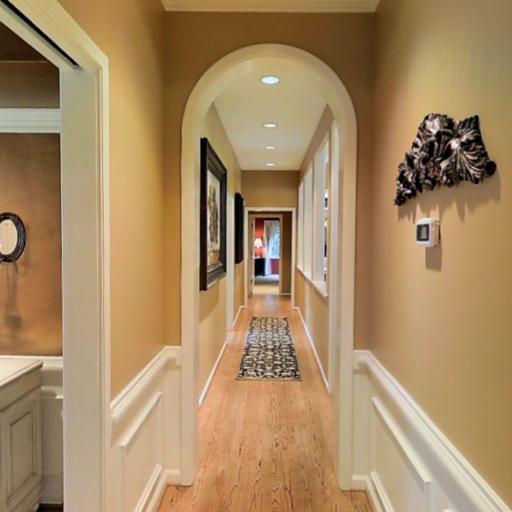}\tskfig{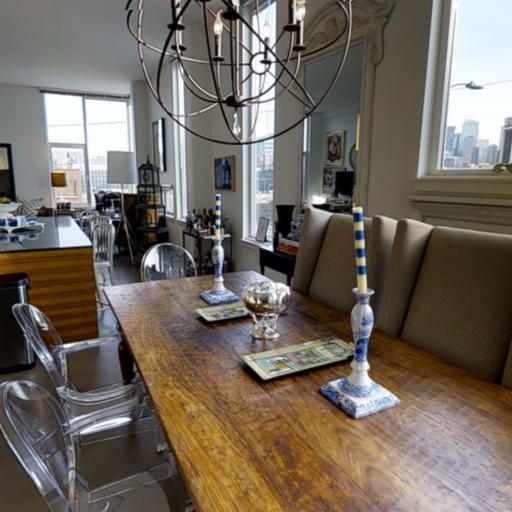}\\
\rotatebox{90}{\hspace{10pt}\footnotesize ours\phantom{g}}&%
\tskfig{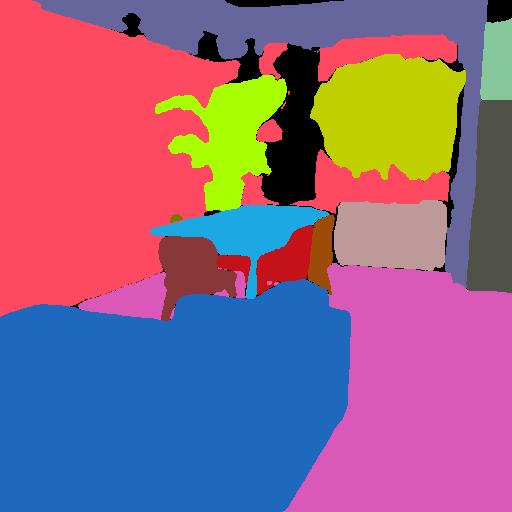}\tskfig{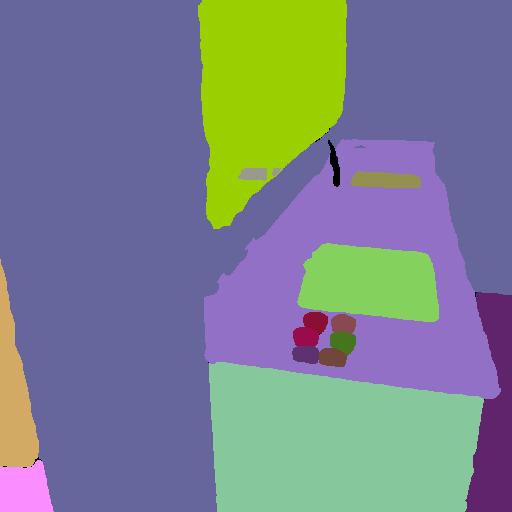}\tskfig{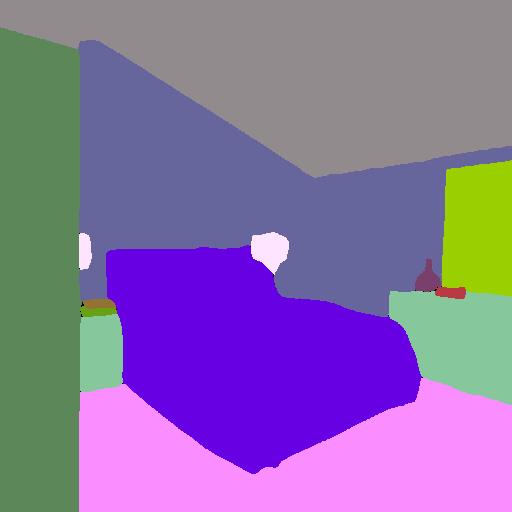}\tskfig{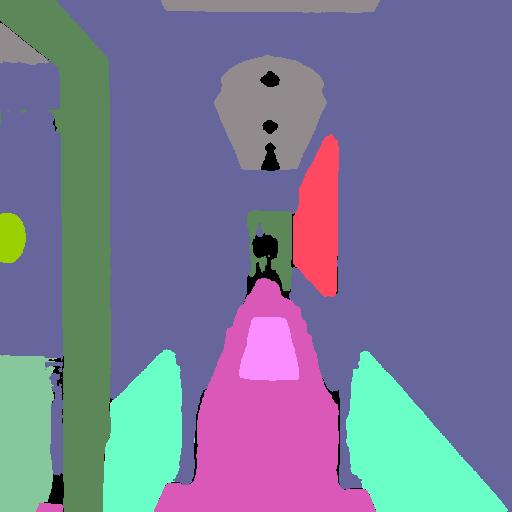}\tskfig{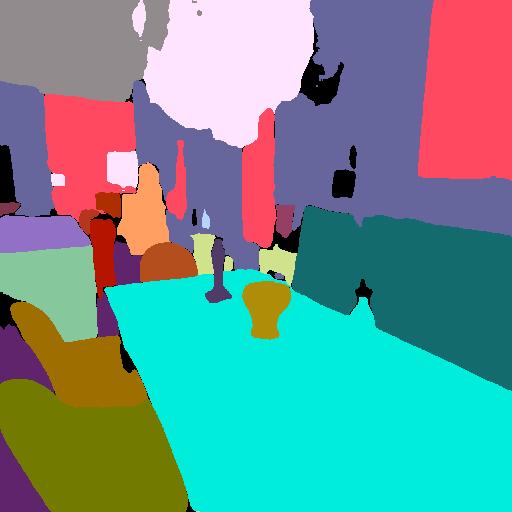}\\
\rotatebox{90}{\hspace{5pt}\footnotesize original}&%
\tskfig{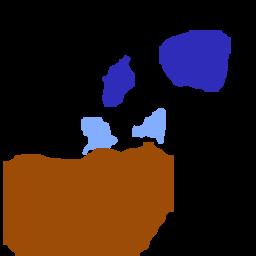}\tskfig{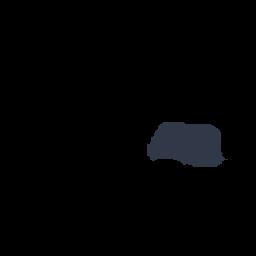}\tskfig{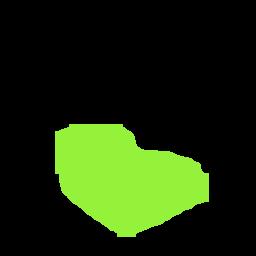}\tskfig{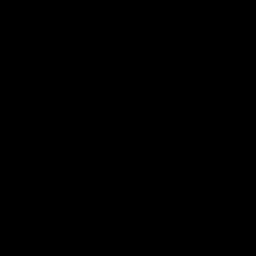}\tskfig{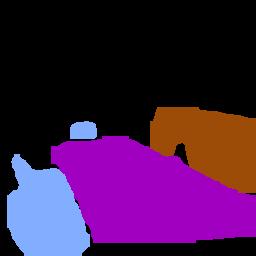}\\
\end{tabular}
\vspace*{-3mm}
\caption{\label{fig:taskonomy}\textbf{Visualization of Taskonomy pseudo-labels: ours (middle) \vs original ones by Li~\etal~\cite{li2017fully} (bottom).} Our pseudo-labels demonstrate higher quality than the existing ones.
}
\end{figure}

To complement the small scale of NYUD-v2 dataset, we also conduct experiments on the Taskonomy~\cite{taskonomy2018} dataset, which is composed of 4.6 million images (train: 3.4M, val: 538K: test: 629K images)
from 537 different buildings with indoor scenes. This dataset is originally developed to facilitate the study of task transfer learning, thereby containing ground-truths of various tasks, including semantic segmentation, depth estimation, surface normal, and so on. The depth Z-buffer and surface normal ground-truth are programmatically computed from image registration and mesh alignment, resulting in high quality annotations. However, the provided semantic segmentation annotations are only pseudo-labels generated by Li et al.~\cite{li2017fully}, an out-dated model trained on COCO~\cite{lin2014microsoft}.
After carefully examining these pseudo-labels, we notice that their quality does not satisfy the requirement for evaluating or improving state-of-the-art models.
Therefore, we adopt the kMaX-DeepLab~\cite{yu2022kmax} model with ConvNeXt-L~\cite{liu2022convnet} backbone pretrained on COCO dataset to regenerate the pseudo-labels for semantic segmentation.
We visually compare both labels in~\figref{fig:taskonomy}.
The Taskonomy dataset enhanced by the new pseudo labels can serve as a meaningful complementary to this research field, particularly given the scarcity of the available large-scale high-quality multi-task dense prediction datasets.
\textit{We will release the high-quality pseudo labels to facilitate future research\footnote{Will release at https://github.com/google-research/deeplab2}}

\textbf{Evaluation Metrics}\quad 
Semantic segmentation is evaluated with mean Intersection-over-Union (mIoU)~\cite{everingham2010pascal}.
For depth estimation, the metrics are root mean square error (RMS), Absolute mean relative error (A.Rel), absolute error in log-scale (Log\textsubscript{10}) and pixel inlier ratio ($\delta_i$) with error threshold at $1.25^i$~\cite{eigen2014depth}.
For surface normal metrics, following~\cite{fouhey2013data,eigen2015predicting} we use mean (Mean) and median (Median) absolute error, RMS angular error, and pixel inlier ratio ($\delta_1$, $\delta_2$, $\delta_3$) with thresholds at 11.5$^{\circ}$, 22.5$^{\circ}$ and 30$^{\circ}$, respectively. 

\textbf{Implementation Details}\quad
We adopt the pixel encoder/decoder and the transformer decoder modules from the kMaX-DeepLab~\cite{yu2022kmax}. Additional L2 normalization is needed at the end of surface normal head to yield 3D unit vectors.
During training, we adopt the same losses from kMaX-DeepLab for semantic segmentation. For depth estimation, when training on NYUD-v2 dataset, we use scale invariant logarithmic error~\cite{eigen2014depth}, relative squared error~\cite{geiger2012we}, following ViP-DeepLab~\cite{qiao2021vip}. We also include the multi-scale gradient loss proposed by MegaDepth~\cite{li2018megadepth} to improve the visual sharpness of the predicted depth map. While training depth estimation on Taskonomy, we switch to robust Charbonnier loss~\cite{barron2019general}, since this dataset has maximum depth value 128m with more outliers. The loss function we apply to surface normal is simply L2 loss, since it has better training stability than truncated angular loss~\cite{do2020surface}.
The experiments on NYUD-v2 are conducted by first pretraining \modelname on Taskonomy dataset, and then finetuning on NYUD-v2 train split. The finetuning step is unnecessary when evaluating on Taskonomy test split. The learning rate at the pretraining and finetuning stages are 5e-4 and 5e-5, respectively. To ensure fair comparisons with previous works, we closely follow their experiment setup. However, due to the use of different training data splits for each of the three tasks in previous studies, we have to train each task independently.

\subsection{Main Results}

\textbf{NYUD-v2}\quad In \tabref{table:main}, we compare \modelname with state-of-the-art models for dense prediction tasks on the NYUD-v2 dataset. We group the existing models based on the number and type of the tasks they support on this dataset. We choose the best numbers reported in the prior works.
As shown in the table, in such a competitive comparison, \modelname still significantly outperforms all the existing models.

Specifically, in the semantic segmentation task, most of the recent models use additional modalities such as depth, but \modelname surpasses the existing best two models CMX~\cite{zhang2023cmx} and CMNeXt~\cite{zhang2023delivering} by $1.2 \%$ mIoU, without using any additional modalities.
In the depth estimation task, the current best model VPD~\cite{zhao2023unleashing} is built upon Stable-Diffusion~\cite{rombach2021highresolution} and pretrained with LAION-5B dataset~\cite{schuhmann2022laion}. Despite of only using a pretraining dataset (\ie, Taskonomy) with $0.1 \%$ of that scale, \modelname achieves better performance on all the depth metrics than VPD.
Furthermore, \modelname breaks the close competition among non-diffusion-based models by a meaningful improvement from above 0.27 to 0.25 RMS (all the recent non-diffusion-based models achieve around 0.27 RMS), similarly for all the other metrics. This is non-trivial, particularly given that the best non-diffusion-based depth model ZoeDepth~\cite{bhat2023zoedepth} uses the pretrained BeiT\textsubscript{384}-L~\cite{beit} backbone and 2-stage training with a mixture of 12 datasets. On the surface normal benchmark, \modelname continues to outperform all the existing models by a substantial margin, with the mean error being reduced from above 14.60 to 13.09.
Finally, when comparing with models that support all three tasks on this dataset, the improvement of \modelname is further amplified for all metrics. Overall, \modelname is not only among the few models that support all the three tasks, but also sets a new state-of-the-art on all three dense prediction benchmarks.

\textbf{Taskonomy}\quad
In~\tabref{table:taskonomy_test}, we report our results on Taskonomy, along with a solid baseline\footnote{We notice recent works~\cite{bhattacharjee2022mult, bhattacharjee2023vision} also report numbers on this benchmark, but with the unreleased code and different experimental setup, we can not consider them here as fair baselines.} DeepLabv3+~\cite{deeplabv3plus2018}. The performance of the same baseline model on NYUD-v2 benchmarks in~\tabref{table:main}, compared with state-of-the-art models, can justify our choice of it as baseline on Taskonomy benchmarks. 
On all three tasks, \modelname significantly outperforms the baseline model. Specifically, the performance on semantic segmentation, depth estimation and surface normal is improved by 8.3 mIoU, 0.13 RMS and 1.0 mean error, respectively. These results further validate the effectiveness of the proposed framework.

\begin{figure*}[t]
\begin{tabular}{p{2.05cm}p{2.05cm}p{2.05cm}p{2.05cm}p{2.05cm}p{2.05cm}p{2.05cm}}{\hspace{20pt}\footnotesize input} & {\hspace{10pt}\footnotesize sem seg pred.} & {\hspace{10pt}\footnotesize ground-truth} & {\hspace{12pt}\footnotesize depth pred.} & {\hspace{10pt}\footnotesize{ground-truth}} & {\hspace{10pt}\footnotesize normal pred.} & {\hspace{10pt}\footnotesize ground-truth}\end{tabular}
\includegraphics[width=0.99\linewidth,trim={0 0 15pt 20pt},clip]{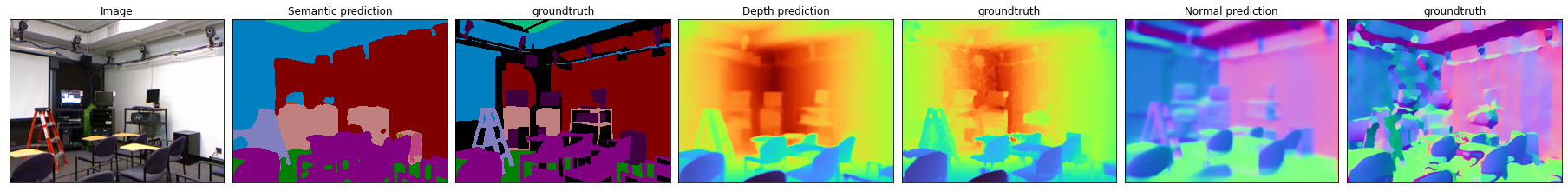}
\includegraphics[width=0.99\linewidth,trim={0 0 15pt 20pt},clip]{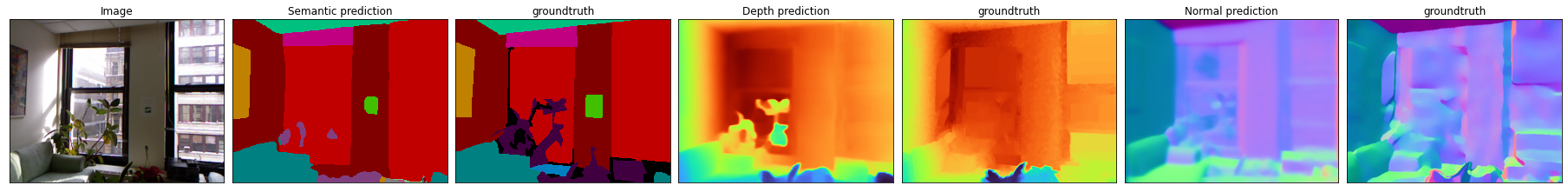}
\vspace*{-3mm}
\caption{
\textbf{Visualization of model inputs and outputs for semantic segmentation, depth estimation and normal prediction.} \modelname resolves fine details on scenes with complex structure (top row), and even performs acceptably when ground-truth is unreliable (last row). 
}
\label{fig:output_visualization}
\end{figure*}

\textbf{Visualizations and Limitations}\quad
We visualize the predictions of \modelname on all the three dense preiction tasks with their corresponding ground-truth in Fig.~\ref{fig:output_visualization}. 
As shown in the top row, \modelname successfully resolves the fine-grained details and irregular object shapes, and predicts high quality results. The bottom row shows a challenging case where \modelname can be further improved. Similar to many other depth models, it is difficult to correctly infer the depth map with glass, mirror and other reflective surfaces, due to the artifacts in the available ground-truth. Lastly, another direction for future improvement is the visual sharpness of the predicted surface normal. We find that the multi-scale gradient loss proposed by MegaDepth~\cite{li2018megadepth} can effectively improve the edge sharpness for depth estimation, and slightly improves depth metrics (\eg, around 0.002 RMS improvement). It is promising to adapt the loss to surface normal prediction to improve visual quality.   

\begin{table*}[t!]
\small
\begin{center}
\scalebox{0.75}{
\begin{tabular}{ c | c c | c | c c c c | c c c c }
 & \multicolumn{2}{c|}{} & \textbf{sem seg} & \multicolumn{4}{c|}{\textbf{depth estimation}} & \multicolumn{4}{c}{\textbf{surface normal}}  \\
 model & encoder & $\text{pretraining data}$ & $\text{mIoU}\uparrow$ & $\text{RMS}\downarrow$ & $\text{A.Rel}\downarrow$ & $\text{Log}_{10}\downarrow$ & $\delta_1\uparrow$ & $\text{Mean}\downarrow$ & $\text{Med}\downarrow$ & $\text{RMS}\downarrow$ & $\delta_1\uparrow$ \\
 \toprule[0.12em]
OMNIVORE~\cite{girdhar2022omnivore} & Swin-L~\cite{liu2021swin} & IN-1K~\cite{russakovsky2015imagenet}, Kinetics~\cite{kay2017kinetics}, SUN~\cite{xiao2016sun} & 56.80  & - & - & - & - & - & - & - & - \\ %
CMX~\cite{zhang2023cmx} & MiT-B5~\cite{xie2021segformer} & IN-1K~\cite{russakovsky2015imagenet} & 56.90 & - & - & - & - & - & - & - & - \\ %
CMNeXt~\cite{zhang2023delivering} & MiT-B4~\cite{xie2021segformer} & IN-1K~\cite{russakovsky2015imagenet} & 56.90  & - & - & - & - & - & - & - & - \\ %
 \hline 
 BinsFormer~\cite{li2022binsformer} & Swin-L~\cite{liu2021swin} & IN-22K~\cite{russakovsky2015imagenet} & - & 0.330 & 0.094 & 0.040 & 0.925 & - & - & - & - \\ %
 DINOv2~\cite{oquab2023dinov2} & ViT-g~\cite{zhai2022scaling} & LVD-142M~\cite{oquab2023dinov2} & - & 0.279 & 0.091 & 0.037 & 0.950 & - & - & - & - \\ %
 AiT-P~\cite{ning2023all} & Swin-L~\cite{liu2021swin} & IN-22K~\cite{russakovsky2015imagenet} & - & 0.275 & 0.076 & 0.033 & 0.954 & - & - & - & - \\ %
 ZoeDepth~\cite{bhat2023zoedepth} & BEiT\textsubscript{384}-L~\cite{beit} & M12~\cite{bhat2023zoedepth} & - & 0.270 & 0.075 & 0.032 & 0.955 & - & - & - & - \\ %
 \hline
 Do et al.~\cite{do2020surface} & ResNeXt-101~\cite{xie2017aggregated} & IN-1K~\cite{russakovsky2015imagenet} & - & - & - & - & - & 16.20 & 8.200 & 25.30 & 59.50  \\ %
 Bae et al.~\cite{bae2021estimating} & MiT-B5~\cite{xie2021segformer} & IN-1K~\cite{russakovsky2015imagenet} & - & - & - & - & - & 14.90 & 7.500 & 23.50 & 62.20  \\ %
 \hline 
MultiMAE~\cite{bachmann2022multimae} & ViT-B~\cite{dosovitskiy2020image} & IN-1K~\cite{russakovsky2015imagenet} & 56.00 & - & - & - & 0.864 & - & - & - & - \\ %
Painter~\cite{wang2023images} & ViT-L~\cite{dosovitskiy2020image} &  IN-1K~\cite{russakovsky2015imagenet} & - & 0.288 & 0.080 & - & 0.950 & - & - & - & -  \\ %
 MTFormer~\cite{xu2022mtformer} & Swin~\cite{liu2021swin} & IN-22K~\cite{russakovsky2015imagenet} & 50.04 & 0.490 & - & - & - & - & - & - & -  \\ %
 \hline

 VPD~\cite{zhao2023unleashing} & SD~\cite{rombach2021highresolution} & LAION-5B~\cite{schuhmann2022laion}, CLIP~\cite{radford2021learning} & - & 0.254 & 0.069 & 0.030 & 0.964 & - & - & - & -  \\ %
 ASNDepth~\cite{long2021adaptive} & HRNet-48~\cite{wang2020deep} & IN-1K~\cite{russakovsky2015imagenet} & - & 0.377 & 0.101 & 0.044 & 0.890 & 20.00 & 13.400 & - & 43.50  \\ %
 iDisc~\cite{piccinelli2023idisc} & Swin-L~\cite{liu2021swin} & IN-22K~\cite{russakovsky2015imagenet} & - & 0.313 & 0.086 & 0.037 & 0.940 & 14.60 & 7.300 & 22.80 & 63.80  \\ %
 \hline
 ARTC~\cite{bruggemann2021exploring} & HRNet-48~\cite{wang2020deep} & IN-1K~\cite{russakovsky2015imagenet} & 46.33 & 0.536 & - & - & - & 20.18 & - & - & -  \\ %
 InvPT~\cite{ye2022invpt} & Swin-L~\cite{liu2021swin} & IN-22K~\cite{russakovsky2015imagenet} & 51.76 & 0.502 & - & - & - & 19.39 & - & - & -  \\ %
 InvPT~\cite{ye2022invpt} & ViT-L~\cite{dosovitskiy2020image} & IN-22K~\cite{russakovsky2015imagenet} & 53.56 & 0.518 & - & - & - & 19.04 & - & - & -  \\ %
 MQTrans.~\cite{xu10multi} & Swin-L~\cite{liu2021swin} & ADE20K~\cite{zhou2017scene} & 54.84 & 0.533 & - & - & - & 19.69 & - & - & -  \\ %
 \hline 
  DeepLabv3+~\cite{deeplabv3plus2018}\dag & ResNet-50~\cite{he2016deep} & IN-1K~\cite{russakovsky2015imagenet}, Taskonomy~\cite{taskonomy2018} & 50.14 & 0.375 & 0.108 & 0.046 & 0.887 & 14.81 & 8.291 & 22.49 & 61.44  \\ %
  \hline\hline
 \modelname & ResNet-50~\cite{he2016deep} & IN-1K~\cite{russakovsky2015imagenet}, Taskonomy~\cite{taskonomy2018} & 51.30 & 0.317 & 0.084 & 0.037 & 0.936 & 14.03 & 7.789 & 21.45 & 63.25  \\ %
  \modelname & ConvNeXt-T~\cite{liu2022convnet} & IN-22K~\cite{russakovsky2015imagenet}, Taskonomy~\cite{taskonomy2018} & 54.59 & 0.282 & 0.076 & 0.033 & 0.954 & 13.73 & 7.533 & 21.18 & 64.07  \\ %
 \modelname & ConvNeXt-S~\cite{liu2022convnet} & IN-22K~\cite{russakovsky2015imagenet}, Taskonomy~\cite{taskonomy2018} & 56.34 & 0.273 & 0.073 & 0.032 & 0.959 & 13.42 & 7.351 & 20.74 & 64.73  \\ %
 \modelname & ConvNeXt-B~\cite{liu2022convnet} & IN-22K~\cite{russakovsky2015imagenet}, Taskonomy~\cite{taskonomy2018} & 57.21 & 0.260 & 0.071 & 0.030 & 0.965 & 13.23 & 7.181 & 20.58 & 65.48  \\ %
 \modelname & ConvNeXt-L~\cite{liu2022convnet} & IN-22K~\cite{russakovsky2015imagenet}, Taskonomy~\cite{taskonomy2018} & \textbf{58.08} & \textbf{0.250} & \textbf{0.067} & \textbf{0.029} & \textbf{0.969} & \textbf{13.09} & \textbf{7.117} & \textbf{20.40} & \textbf{65.66}   %
\end{tabular}
}
\end{center}
\vspace*{-6mm}
\caption{
\textbf{NYUD-v2 \textit{test} set results.}
Our \modelname with ConvNeXt-L consistently outperforms prior arts on three benchmarks of NYUD-v2. \dag: Our reimplemented DeepLabv3+ of the pixel-regression method for depth estimation and surface normal prediction serves as an additional strong baseline.
} %
\label{table:main}
\end{table*}

\begin{table*}[t!]
\begin{center}
\scalebox{0.7}{
\begin{tabular}{ c | c | c | c c c c | c c c c }
 &  & \textbf{sem seg} & \multicolumn{4}{c|}{\textbf{depth estimation}} & \multicolumn{4}{c}{\textbf{surface normal}}  \\
 model & encoder & $\text{mIoU}\uparrow$ & $\text{RMS}\downarrow$ & $\text{A.Rel}\downarrow$ & $\text{Log}_{10}\downarrow$ & $\delta_1\uparrow$ & $\text{Mean}\downarrow$ & $\text{Med}\downarrow$ & $\text{RMS}\downarrow$ & $\delta_1\uparrow$ \\
 \toprule[0.12em]
 DeepLabv3+~\cite{deeplabv3plus2018} & ResNet-50~\cite{he2016deep} & 32.87 & 0.6303 & 0.1472 & 0.0618 & 82.61 & 11.45 & 4.938 & 19.65 & 56.13  \\ %
 \hline \hline
 \modelname & ResNet-50~\cite{he2016deep}  & 35.00 & 0.5885 & 0.1236 & 0.0538 & 86.49 & 10.96 & 4.922 & 18.83 & 55.27  \\ %
  \modelname & ConvNeXt-T~\cite{liu2022convnet}  & 35.23 & 0.5287 & 0.1140 & 0.0479 & 88.56 & 10.87 & 4.859 & 18.49 & 55.39  \\ %
 \modelname & ConvNeXt-S~\cite{liu2022convnet}  & 38.99 & 0.5097 & 0.1077 & 0.0452 & 89.59 & 10.67 & 4.709 & 18.46 & 56.05  \\ %
 \modelname & ConvNeXt-B~\cite{liu2022convnet}  & \textbf{41.17} & \textbf{0.4981} & 0.1034 & 0.0433 & 90.37 & 10.58 & 4.646 & 18.37 & 56.43  \\ %
 \modelname & ConvNeXt-L~\cite{liu2022convnet}  & 40.94 & 0.4988 & \textbf{0.1029} & \textbf{0.0432} & \textbf{90.41} & \textbf{10.46} & \textbf{4.550} & \textbf{18.25} & \textbf{56.76}  %
\end{tabular}
}
\end{center}
\vspace*{-6mm}
\caption{
\textbf{Taskonomy \textit{test} set results.}
The DeepLabv3+ baseline and \modelname are trained with the same configuration.
} 
\label{table:taskonomy_test}
\end{table*}

\subsection{Ablation Studies}

\begin{figure*}[t]
\begin{subfigure}[b]{0.32\textwidth}
\includegraphics[width=\linewidth]{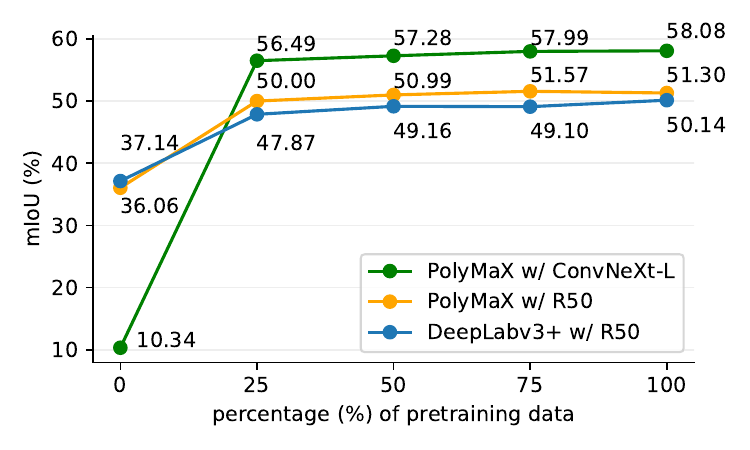}
\caption{\label{fig:semseg_scalability} Scalability of semantic segmentation model   
}
\end{subfigure}
\hfill
\begin{subfigure}[b]{0.32\textwidth}
\includegraphics[width=\linewidth]{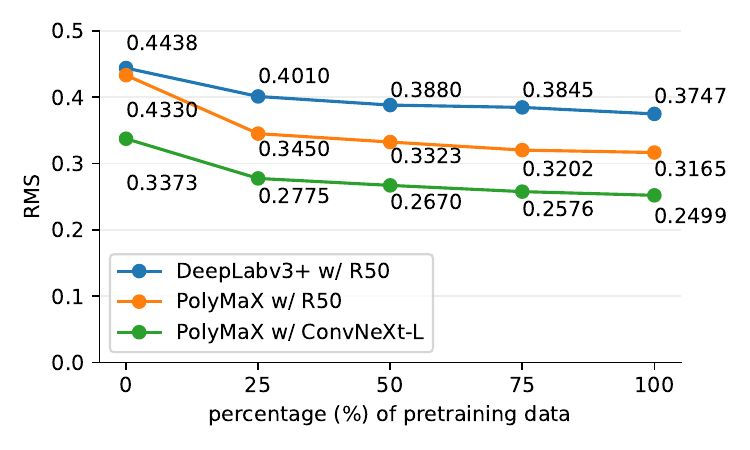}
\caption{
\label{fig:depth_scalability} Scalability of depth estimation model
}
\end{subfigure}
\hfill
\begin{subfigure}[b]{0.32\textwidth}
\includegraphics[width=\linewidth]{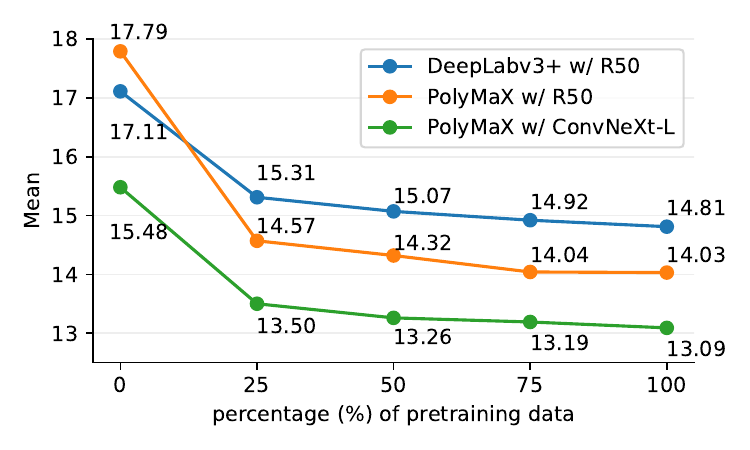}
\caption{\label{fig:normal_scalability} Scalability of surface normal prediction model}
\end{subfigure}
\hfill
\vspace*{-3mm}
\caption{\label{fig:scalability}
\textbf{Experiments of model scalability with different pretraining data percentage of Taskonomy on NYUD-v2.}
Our \modelname demonstrates better scalability than the solid per-pixel prediction baseline DeepLabv3+ on all three tasks.
}
\end{figure*}

\begin{figure}[t]
\centering
\includegraphics[width=0.9\linewidth]{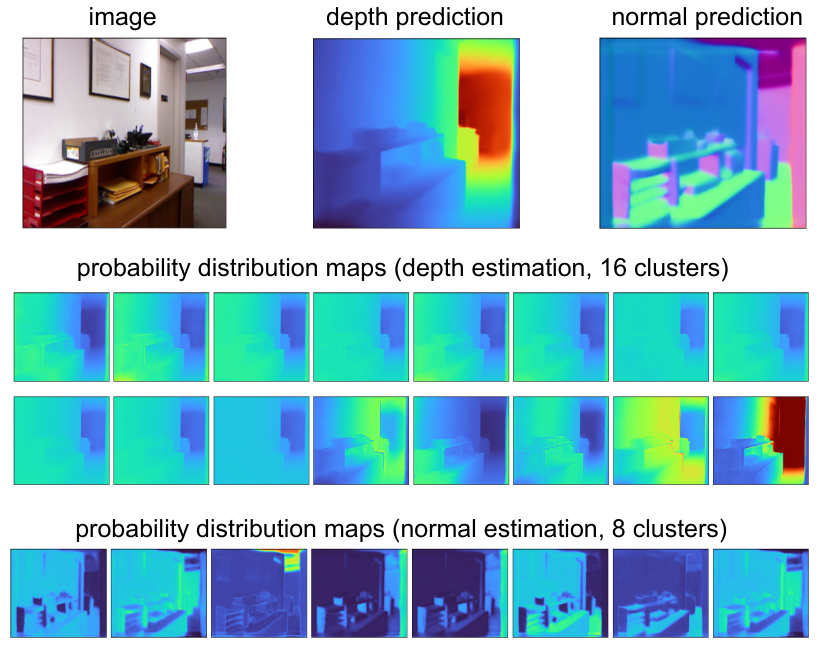}
\vspace*{-3mm}
\caption{\textbf{Visualization of probability distribution maps} for depth prediction (middle) and normal estimation (bottom) tasks. It illustrates the attention on regions with different depth and different angles, which validate the effectiveness of \modelname with cluster-prediction paradigm. 
}
\label{fig:cluster-viz}
\end{figure}

\textbf{Scalability of Mask-Transformer-Based Framework}\quad
We investigate the scalability of \modelname, since this property has become increasingly crucial in the large scale model era. Fig.~\ref{fig:scalability} demonstrates that \modelname's scalability with the percentage of pretraining dataset. For a fair comparison, we build the baseline models upon DeepLabv3+~\cite{deeplabv3plus2018}, using pixel-classification prediction head for semantic segmentation, and pixel-regression prediction head for depth estimation and surface normal prediction. The training configurations are the same within each task, except that the DeepLabv3+ based depth model requires the robust Charbonnier loss~\cite{barron2019general} to overcome the typical training instability issue of regression with outliers.   

As shown in Fig.~\ref{fig:semseg_scalability}, \modelname scales significantly better than the baseline model on semantic segmentation. When all models are pretrained with only ImageNet~\cite{russakovsky2015imagenet} (\ie, 0\% Taskonomy data is introduced in pretraining), \modelname performs worse than the baseline, especially the ConvNeXt-L model variant fails disappointingly. This is because the larger capacity of \modelname can not be fully exploited with only 795 training samples that are available for semantic segmentation from NYUD-v2 dataset. Thus, after including only $25 \%$ of the Taskonomy data in pretraining, \modelname with ResNet-50 and ConvNeXt-L immediately outperform the baseline model with more than $2 \%$ mIoU and $9 \%$ mIoU, respectively. This performance gaps remain when the percentage of pretraining data increases, and stay at more than $1 \%$ and $7 \%$ mIoU after using the entire Taskonomy dataset for pretraining. 

The scalability of depth estimation is shown in Fig.~\ref{fig:depth_scalability}. When no Taskonomy pretraining data is used, the improvement of \modelname over the baseline (both with ResNet-50) is only 0.01 RMS, implying negligible positive impact of cluster-prediction paradigm. The more than 0.1 RMS performance improvement of \modelname with ConvNeXt-L mainly comes from the more powerful backbone. However, \modelname has better scalability with the increase of pretraining data. When $25 \%$ Taskonomy data is involved in pretraining, the model performance improvements of \modelname with ResNet-50 and ConvNeXt-L increase from 0.01 and 0.1 RMS to 0.05 and 0.12 RMS, respectively. With all Taskonomy dataset being used in pretraining, the performance gaps of the two \modelname come to 0.6 and 0.12 RMS. 

The scalability of \modelname on surface normal, showing a similar trend as depth estimation, is illustrated in Fig.~\ref{fig:normal_scalability}. Even though \modelname with ResNet-50 does not perform better than the baseline model when no Tasknomy pretraining data is used, it significantly surpasses the baseline model by 0.7 and 0.9 mean error at $25 \%$ and $100 \%$ of pretraining data. While \modelname with ConvNeXt-L substantially outperforms the baseline model by 1.2 and 1.8 mean error at $25 \%$ and $100 \%$ of pretraining data.

\paragraph{Visualization of Cluster Representation}
To obtain deeper understanding of the model behaviour, we further look into the visualization of the learned cluster centers and the probability distribution across those clusters for depth estimation and surface normal, shown in Fig.~\ref{fig:cluster-viz}. Interestingly, 12 out of the 16 probability maps from the depth estimation model look similar, implying redundancy in the learned 16 cluster centers. Similar findings are observed for surface normal.
This can also explain the small impact of cluster count $K$ on model performance in tables of supplementary.

Despite the redundancy, the rest unique probability maps clearly depict the cluster-prediction paradigm. For instance, the last 4 probability maps of the depth model correspond to the attention on close, medium, futher and furthest regions in the scene. Similarly, the unique probability maps of surface normal illustrate the attention on surfaces with different angles, including left, right, upwards, downwards.

\section{Conclusion}
\label{sec:conclusion}
We have generalized dense prediction tasks with the same mask transformer framework, realized by casting semantic segmentation, depth estimation and surface normal to cluster-prediction paradigm.   
The proposed \modelname has demonstrated state-of-the-art results on the three benchmarks of NYUD-v2 dataset.
We hope the superior performance of \modelname can enable many impactful applications, including but not limited to scene understanding, image generation and editing, and augmented reality.

\textit{Acknowledgement} We thank Yukun Zhu, Jun Xie and Shuyang Sun for their support on the code-base.

{\small
\bibliographystyle{ieee_fullname}
\bibliography{egbib}
}

\clearpage
\appendix

\section*{Supplementary Materials}
\label{sec:sup}

In the supplementary materials, we provide additional information as listed below:

\begin{itemize}
    \item \secref{sec:training} provides detailed training protocol used in the experiments.
    \item \secref{sec:cluster} provides additional ablations studies.
    \item \secref{sec:viz} provides more visualizations of (1) model predictions, (2) failure modes, (3) learned probability distribution maps, and (4) our generated high-quality pseudo-labels for Taskonomy semantic segmentation.
\end{itemize}

\section{Training Protocol}
\label{sec:training}
The training configurations of \modelname closely follow kMaX-DeepLab, including the regularization, drop path~\cite{huang2016deep}, color jitting~\cite{cubuk2018autoaugment}, AdamW optimizer~\cite{kingma2014adam,loshchilov2017decoupled} with weight decay 0.05, and learning rate multiplier 0.1 for backbone. Additionally, for depth estimation and surface normal, we follow the data preprocessing in~\cite{li2022binsformer}, except that we disable random scaling and rotation for surface normal. 

\section{Additional Ablation Studies}
\label{sec:cluster}
\textbf{Impact of Cluster Granularity}\quad
We analyze the impact of cluster granularity (\ie, $K$ cluster centers) for depth estimation and surface normal, which are presented in Tab.~\ref{table:depth_cluster} and Tab.~\ref{table:normal_cluster}. Note that, we skip this analysis for semantic segmentation, as we can simply assign the number of clusters as the number of classes. In both Tab.~\ref{table:depth_cluster} and Tab.~\ref{table:normal_cluster}, we observe that the cluster granularity does not have a significant impact on the model performance on either benchmarks. Among the different cluster settings, 16 clusters and 8 clusters perform the best for depth estimation and for surface normal, respectively.  

\begin{table}[ht!]
\scalebox{0.9}{
\begin{tabular}{c||ccc|ccc}
$K$ & $\text{RMS}\downarrow$ & $\text{A.Rel}\downarrow$ & $\text{Log}_{10}\downarrow$ & $\delta_1\uparrow$ & $\delta_2\uparrow$  & $\delta_3\uparrow$   \\
\toprule[0.12em]
 4 & 0.2544 & 0.0689 & 0.0295 & 96.65 & 99.53 & \textbf{99.91} \\ 
 8 & 0.2578 & 0.0691 & 0.0296 & 96.64 & 99.58 & 99.89 \\  
 16 & \textbf{0.2499} & \textbf{0.0670} & \textbf{0.0288} & \textbf{96.90} & 99.58 & 99.90 \\  
 32 & 0.2520 & 0.0685 & 0.0293 & 96.44 & 99.56 & \textbf{99.91} \\  
 64 & 0.2537 & 0.0688 & 0.0295 & 96.77 & \textbf{99.61} & 99.90 \\  
\end{tabular}
}
\caption{\label{table:depth_cluster}
\textbf{Impact of number of clusters ($K$) on depth estimation.}} 
\end{table}

\begin{table}[ht!]
\scalebox{0.9}{
\begin{tabular}{c||ccc|ccc}
$K$ & $\text{Mean}\downarrow$ & $\text{Med}\downarrow$ & $\text{RMS}\downarrow$ & $\delta_1\uparrow$ & $\delta_2\uparrow$  & $\delta_3\uparrow$   \\ 
\toprule[0.12em]
 4 & 13.10 & \textbf{7.075} & 0.2046 & 65.74 & 82.19 & 87.74 \\  
 8 & \textbf{13.09} & 7.117 & \textbf{0.2040} & 65.66 & \textbf{82.28} & \textbf{87.83} \\  
 16 & 13.15 & 7.111 & 0.2051 & 65.70 & 82.17 & 87.73 \\  
 32 & 13.11 & \textbf{7.075} & 0.2048 & \textbf{65.75} & 82.23 & 87.77 \\ 
\end{tabular}
}
\caption{\label{table:normal_cluster}
\textbf{Impact of number of clusters ($K$) on surface normal.}
} 
\end{table}

\section{Additional Visualization}
\label{sec:viz}

\begin{figure*}[t]
\begin{tabular}{p{2.05cm}p{2.05cm}p{2.05cm}p{2.05cm}p{2.05cm}p{2.05cm}p{2.05cm}}{\hspace{20pt}\footnotesize input} & {\hspace{10pt}\footnotesize sem seg pred.} & {\hspace{10pt}\footnotesize ground-truth} & {\hspace{12pt}\footnotesize depth pred.} & {\hspace{10pt}\footnotesize{ground-truth}} & {\hspace{10pt}\footnotesize normal pred.} & {\hspace{10pt}\footnotesize ground-truth}\end{tabular}
\includegraphics[width=0.99\linewidth,trim={0 0 15pt 20pt},clip]{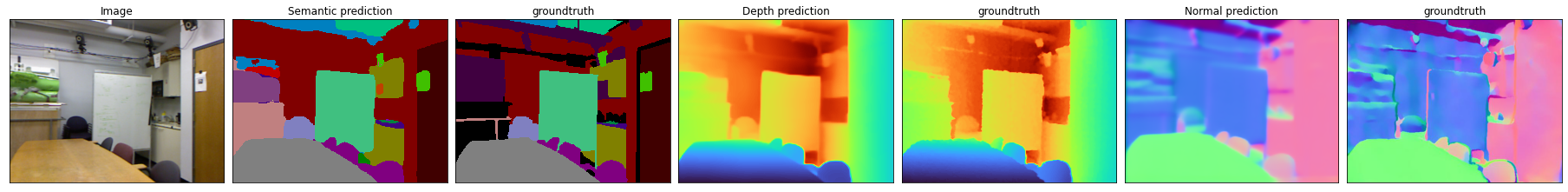}
\includegraphics[width=0.99\linewidth,trim={0 0 15pt 20pt},clip]{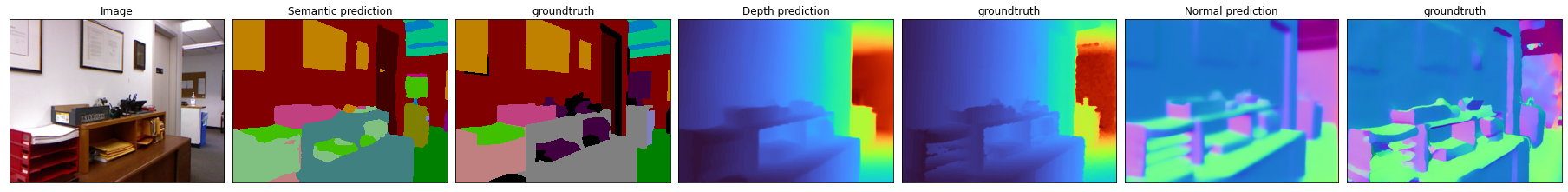}
\includegraphics[width=0.99\linewidth,trim={0 0 15pt 20pt},clip]{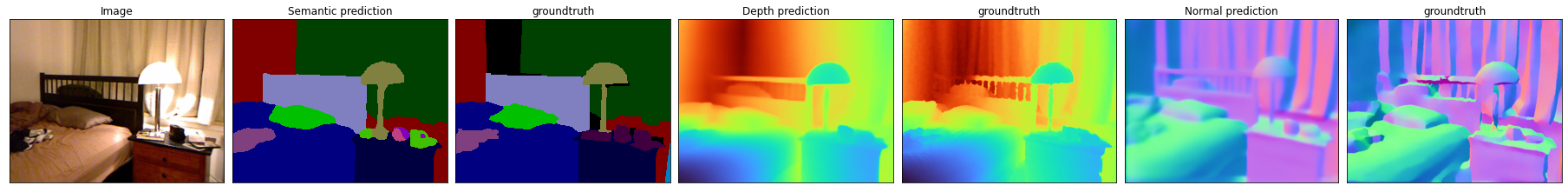}
\includegraphics[width=0.99\linewidth,trim={0 0 15pt 20pt},clip]{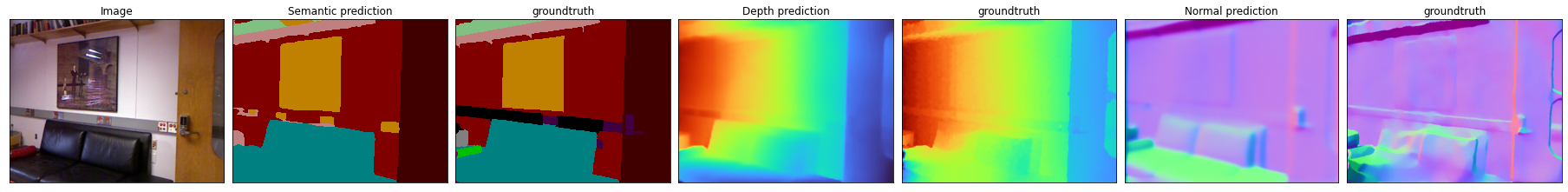}

\includegraphics[width=0.99\linewidth,trim={0 0 15pt 20pt},clip]{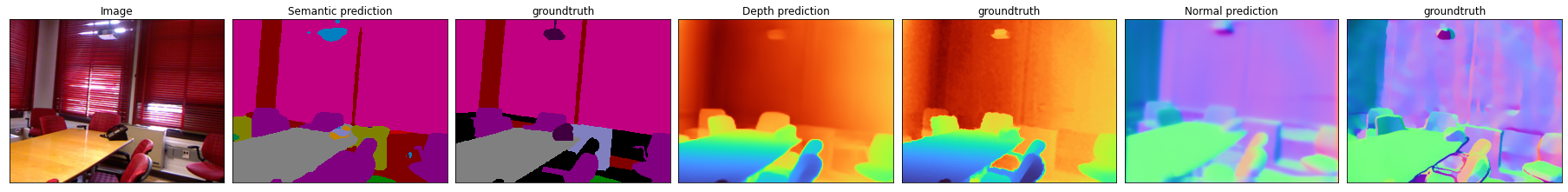}
\includegraphics[width=0.99\linewidth,trim={0 0 15pt 20pt},clip]{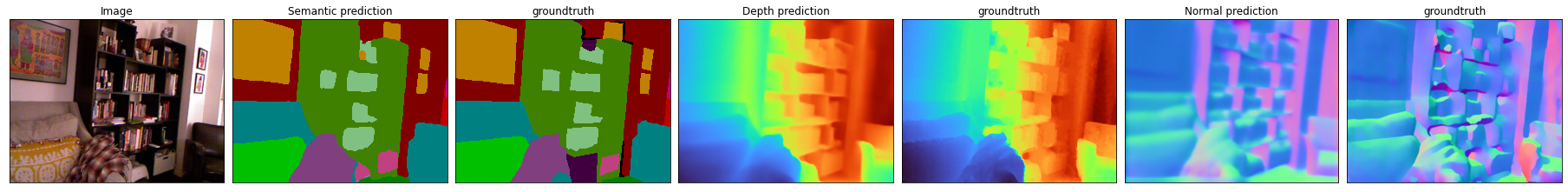}
\includegraphics[width=0.99\linewidth,trim={0 0 15pt 20pt},clip]{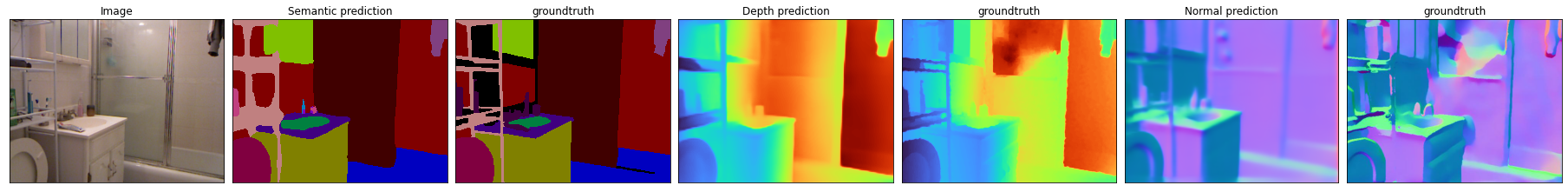}
\caption{
\textbf{Visualization of model inputs and outputs for semantic segmentation, depth estimation and normal prediction.} \modelname is capable of capturing fine details on scenes with complex structures. Interestingly, as shown in the bottom row, \modelname can even reasonably estimate the depth for the glass door, where depth models typically struggle. 
}
\label{fig:sup_output_visualization}

\end{figure*}

\textbf{Model Predictions}\quad
In~\figref{fig:sup_output_visualization}, we show more model predictions of semantic segmentation, depth estimation, and surface normal prediction. As shown in the figure, our proposed \modelname can capture fine details on scenes with complex structures.

\begin{figure*}[t]
\begin{tabular}{p{2.05cm}p{2.05cm}p{2.05cm}p{2.05cm}p{2.05cm}p{2.05cm}p{2.05cm}}{\hspace{20pt}\footnotesize input} & {\hspace{10pt}\footnotesize sem seg pred.} & {\hspace{10pt}\footnotesize ground-truth} & {\hspace{12pt}\footnotesize depth pred.} & {\hspace{10pt}\footnotesize{ground-truth}} & {\hspace{10pt}\footnotesize normal pred.} & {\hspace{10pt}\footnotesize ground-truth}\end{tabular}
\includegraphics[width=0.99\linewidth,trim={0 0 15pt 20pt},clip]{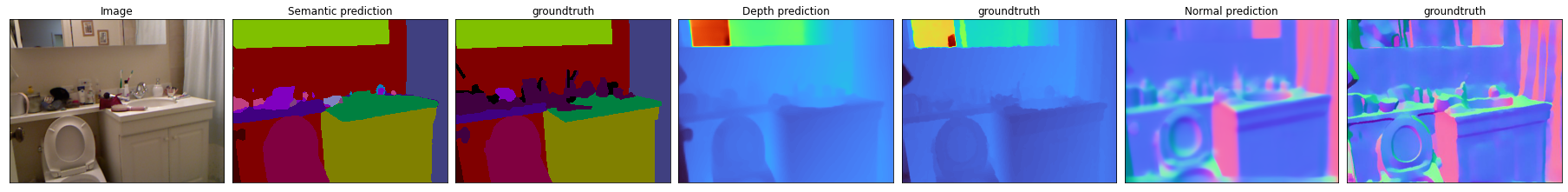}
\includegraphics[width=0.99\linewidth,trim={0 0 15pt 20pt},clip]{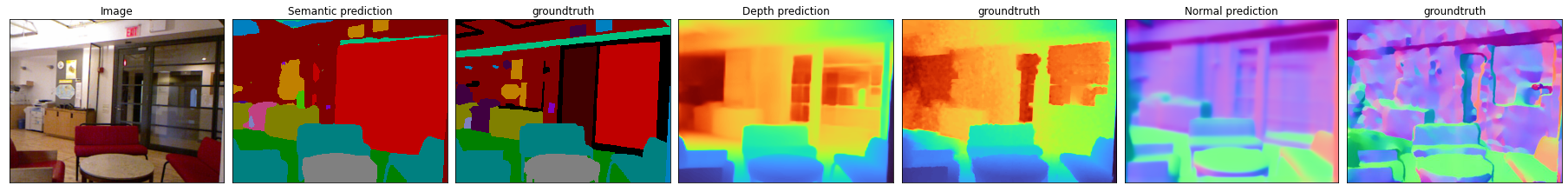}

\caption{
[\textbf{Failure mode}] \modelname still has difficulties with correctly predicting the depth and surface normal for transparent and reflective surfaces (e.g. mirror in first row, glass in second row). These are well-known challenges for such tasks, especially the ground-truths in these scenarios are also often unreliable, as shown in these examples.
}
\label{fig:sup_output_failure1}
\end{figure*}

\begin{figure*}[t]
\begin{tabular}{p{2.05cm}p{2.05cm}p{2.05cm}p{2.05cm}p{2.05cm}p{2.05cm}p{2.05cm}}{\hspace{20pt}\footnotesize input} & {\hspace{10pt}\footnotesize sem seg pred.} & {\hspace{10pt}\footnotesize ground-truth} & {\hspace{12pt}\footnotesize depth pred.} & {\hspace{10pt}\footnotesize{ground-truth}} & {\hspace{10pt}\footnotesize normal pred.} & {\hspace{10pt}\footnotesize ground-truth}\end{tabular}
\includegraphics[width=0.99\linewidth,trim={0 0 15pt 20pt},clip]{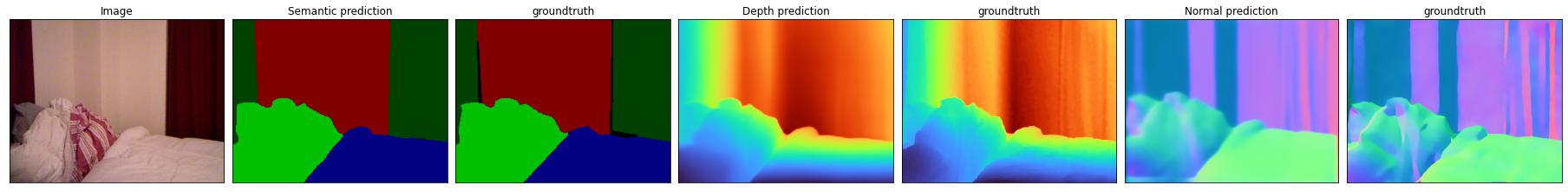}
\includegraphics[width=0.99\linewidth,trim={0 0 15pt 20pt},clip]{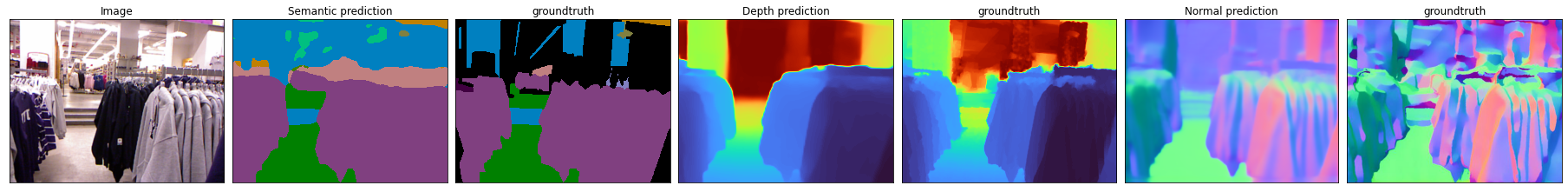}
\caption{
[\textbf{Failure mode}] Although \modelname achieves superior performance on all three benchmarks on NYUD-v2 dataset, we observe that it still suffers from the over-smoothness issue for depth estimation and surface normal tasks, which other prior works~\cite{bae2021estimating,Ranftl2022} attempt to tackle.
}
\label{fig:sup_output_failure2}
\end{figure*}

\textbf{Failure Modes}\quad
To better understand the limitations of the proposed model, we also look into the failure modes.
As shown in~\figref{fig:sup_output_failure1}, \modelname struggles to predict the depth and surface normal for transparent and reflective objects, which are the most challenging issues in the tasks of depth and surface normal estimation. The difficulties can also be reflected by the unreliable ground-truth annotations for those cases. In~\figref{fig:sup_output_failure2}, our model sometimes predicts over-smoothed depth and surface normal results. The findings of~\cite{bae2021estimating,Ranftl2022} (\eg, a better loss function) may alleviate this issue, which is left for future exploration.

\begin{figure*}[t]
\centering
\includegraphics[width=0.99\linewidth]{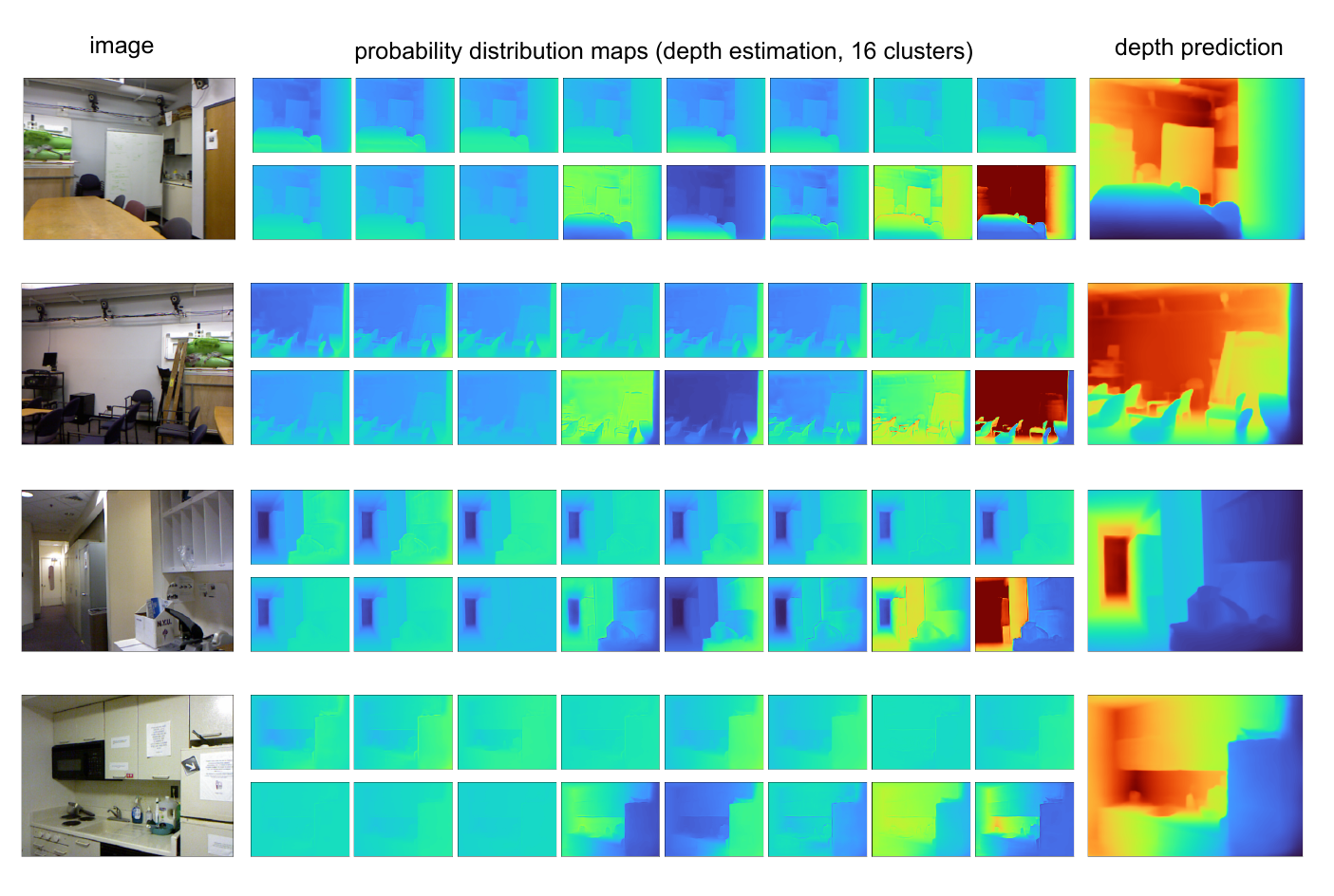}
\vspace*{-3mm}
\caption{\textbf{Additional visualization of probability distribution maps for depth estimation}. Despite of the redundancy in the 16 probability distribution maps, the unique ones clearly demonstrate that the pixels are clustered as closest, mid-range, and furthest distances, which validate the effectiveness of \modelname with cluster-prediction paradigm.
}
\label{fig:sup_cluster-viz_1}
\end{figure*}

\begin{figure*}[t]
\centering
\includegraphics[width=0.99\linewidth]{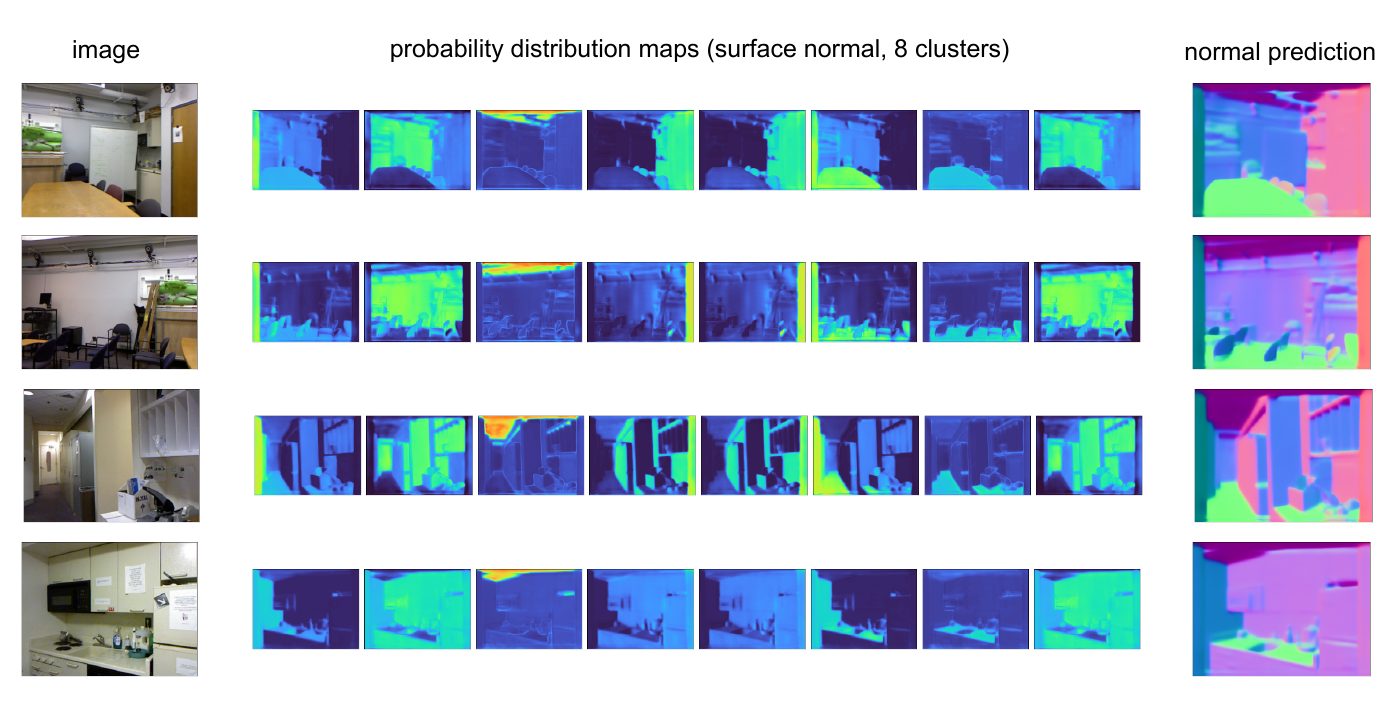}
\vspace*{-3mm}
\caption{\textbf{Additional visualization of probability distribution maps for surface normal prediction}. These probability maps highlight regions with different angles, demonstrating \modelname is capable of clustering pixels based on the normal directions.
}
\label{fig:sup_cluster-viz_2}
\end{figure*}

\textbf{Probability Distribution Maps}\quad
We provide additional visualizations of the learned probability distribution maps for depth estimation and surface normal prediction in~\figref{fig:sup_cluster-viz_1} and~\figref{fig:sup_cluster-viz_2}, respectively.
As shown in the figures, the learned probability distribution maps effectively cluster pixels for different distances (for depth task) or angles (for surface normal task).

\begin{figure*}[t]
\centering
\newcommand{\tskfig}[1]{\includegraphics[width=0.15\linewidth]{figs/sup_figs/sup-taskonomy/#1.jpeg}\hspace{2pt}}
\begin{tabular}{r@{\hspace{2pt}}l}
\rotatebox{90}{\hspace{10pt}\footnotesize input}&%
\tskfig{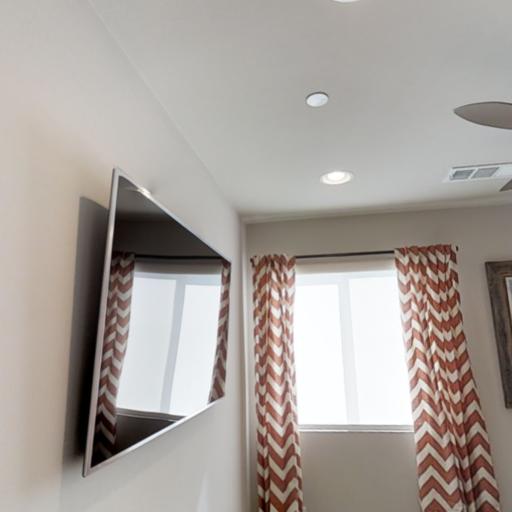}\tskfig{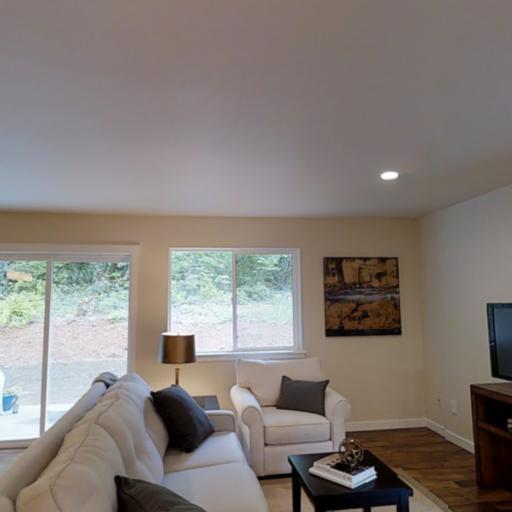}\tskfig{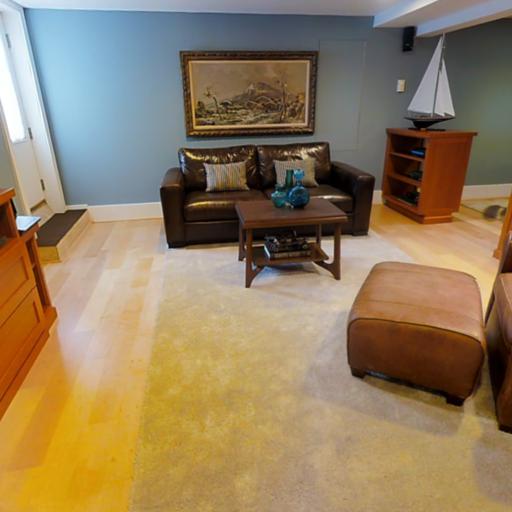}\tskfig{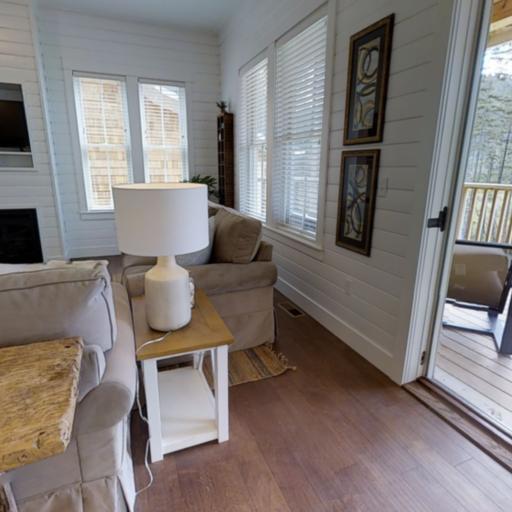}\tskfig{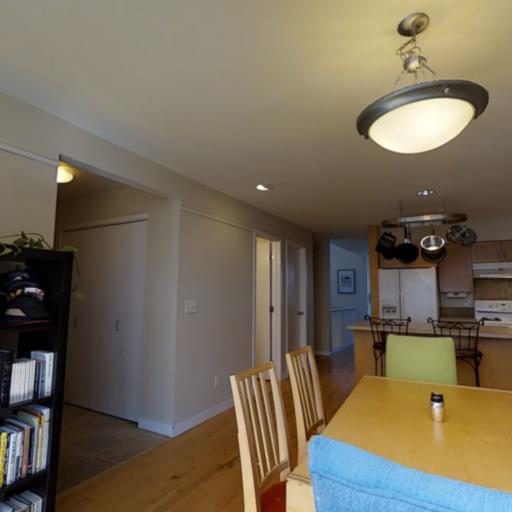}\tskfig{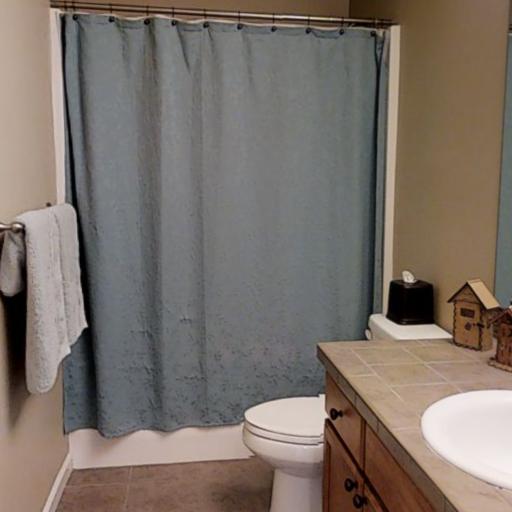}\\
\rotatebox{90}{\hspace{10pt}\footnotesize ours\phantom{g}}&%
\tskfig{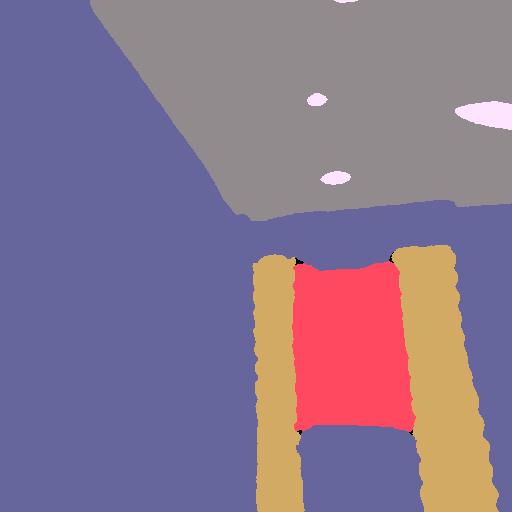}\tskfig{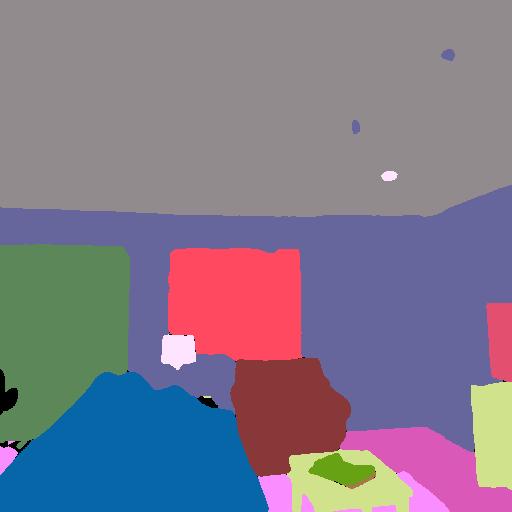}\tskfig{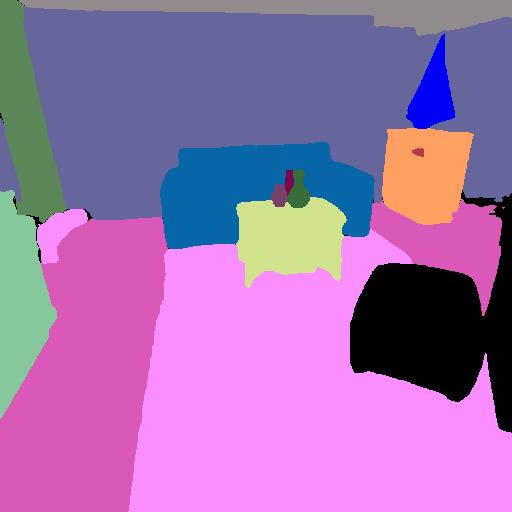}\tskfig{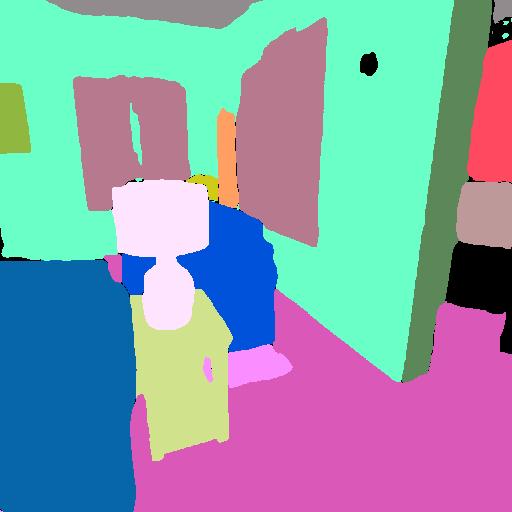}\tskfig{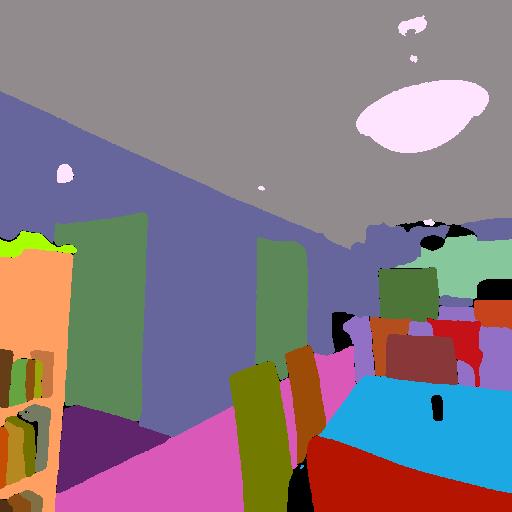}\tskfig{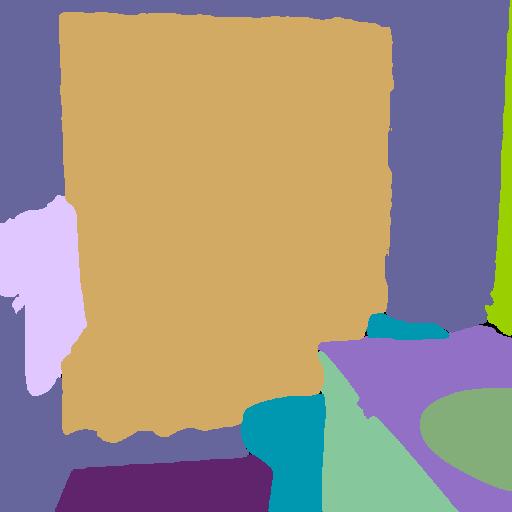}\\
\rotatebox{90}{\hspace{5pt}\footnotesize original}&%
\tskfig{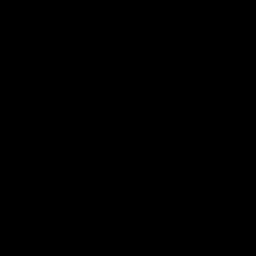}\tskfig{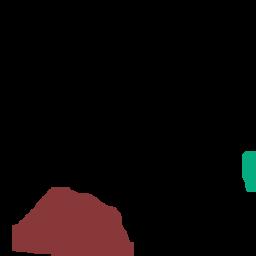}\tskfig{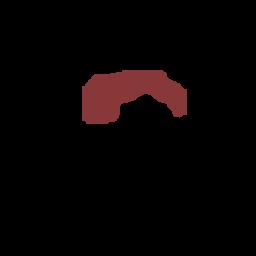}\tskfig{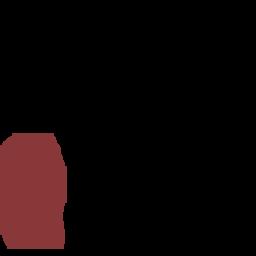}\tskfig{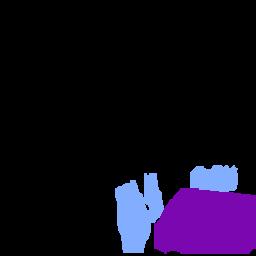}\tskfig{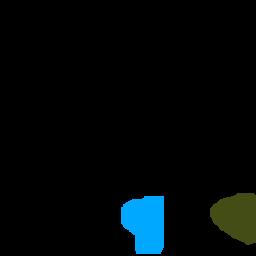}\\
\end{tabular}
\caption{\label{fig:sup_taskonomy}\textbf{Additional visualization of Taskonomy pseudo-labels: ours (middle) \vs original ones by Li~\etal~\cite{li2017fully} (bottom).} Our pseudo-labels demonstrate higher quality than the existing ones.
}
\end{figure*}

\textbf{Taskonomy Pseudo-Labels}\quad
In~\figref{fig:sup_taskonomy}, we show additional visualization of the generated high-quality pseudo-labels for Taskonomy semantic segmentation.

\end{document}